\documentclass[journal]{IEEEtran}
\usepackage{multirow} 
\usepackage{bm}
\usepackage{times}
\usepackage{soul}
\usepackage{url}
\usepackage[hidelinks]{hyperref}
\usepackage[utf8]{inputenc}
\usepackage[small]{caption}
\usepackage{subcaption}
\usepackage{graphicx}
\usepackage{amsmath}
\usepackage{amssymb}
\usepackage{amsthm}
\usepackage{booktabs}
\usepackage{algorithm}
\usepackage{algorithmic}
\usepackage{xcolor}  
\definecolor{Green}{rgb}{0,0.5,0}
\definecolor{steelblue}{RGB}{70,130,180}
\usepackage[switch]{lineno}
\usepackage{pifont}
\DeclareMathOperator*{\argmin}{argmin}
\begin{document}

\title{Mitigating Domain Shift in Federated Learning via Intra- and Inter-Domain Prototypes}
\author{~Huy~Q.~Le, Ye Lin Tun, Yu Qiao,~Minh~N.~H.~Nguyen,\IEEEmembership{~Member,~IEEE}, Keon Oh Kim, \\ Eui-Nam Huh,\IEEEmembership{~Member,~IEEE} and Choong Seon Hong,\IEEEmembership{~Fellow,~IEEE}.

        \IEEEcompsocitemizethanks{	
           \IEEEcompsocthanksitem Huy~Q.~Le, Ye Lin Tun, Keon Oh Kim, Eui-Nam Huh,~and~Choong~Seon~Hong are with the Department of Computer Science and Engineering, School of Computing, Kyung Hee University, Yongin-si 17104, South Korea. (email: \{quanghuy69, yelintun, keonoh, johnhuh, cshong\}@khu.ac.kr).
            \IEEEcompsocthanksitem Yu~Qiao is with the Department of Artificial Intelligence, School of Computing, Kyung Hee University, Yongin-si 17104, South Korea. (email: qiaoyu@khu.ac.kr).
           \IEEEcompsocthanksitem M.~N.~H.~Nguyen is with Digital Science and Technology Institute, The University of Danang—Vietnam-Korea University of Information and Communication Technology, Da Nang, Vietnam (email: nhnminh@vku.udn.vn).
                                                        }
        }



\maketitle

\begin{abstract}
Federated Learning (FL) has emerged as a decentralized machine learning technique, allowing clients to train a global model collaboratively without sharing private data. However, most FL studies ignore the crucial challenge of heterogeneous domains where each client has a distinct feature distribution, which is popular in real-world scenarios. Prototype learning, which leverages the mean feature vectors within the same classes, has become a prominent solution for federated learning under domain shift. However, existing federated prototype learning methods focus soley on inter-domain prototypes and neglect intra-domain perspectives. In this work, we introduce a novel federated prototype learning method, namely I$^2$PFL, which incorporates \textbf{I}ntra-domain and \textbf{I}nter-domain \textbf{P}rototypes, to mitigate domain shift from both perspectives and learn a generalized global model across multiple domains in federated learning. To construct intra-domain prototypes, we propose feature alignment with MixUp-based augmented prototypes to capture the diversity within local domains and enhance the generalization of local features. Additionally, we introduce a reweighting mechanism for inter-domain prototypes to generate generalized prototypes that reduce domain shift while providing inter-domain knowledge across multiple clients. Extensive experiments on the Digits, Office-10, and PACS datasets illustrate the superior performance of our method compared to other baselines.
\end{abstract}


\section{Introduction}



Federated Learning (FL) has emerged as a prominent distributed machine learning framework, enabling multiple clients to collaboratively train a model without leaking private data~\cite{mcmahan2017communication,li2020federated,li2021survey}. The widely used FL approach, FedAvg~\cite{mcmahan2017communication}, ensures user privacy by sharing only model parameters with a central server. In recent years, FL has gained considerable attention and demonstrated promising results across various domains~\cite{rieke2020future,kairouz2021advances,xu2023federated,liu2024vertical,chen2024think,wang2025federated_tmm}. Despite its potential, FL faces a critical challenge: data heterogeneity~\cite{li2020federated,mendieta2022local,li2022federated}. As such, the data distributions across clients are non-independently and identically distributed, i.e., non-iid, leading to the degradation of learning performance and fluctuations in the convergence of the global model performance.
To address this challenge, recent FL methods have focused on enhancing local training through regularization techniques~\cite{li2020federated,pmlr-v119-karimireddy20a} or novel aggregation schemes~\cite {hsu2019measuring,Wang2020Federated,wang2024aggregation}. 
\begin{figure}[]
	\centering
	\includegraphics[width=0.95\linewidth]{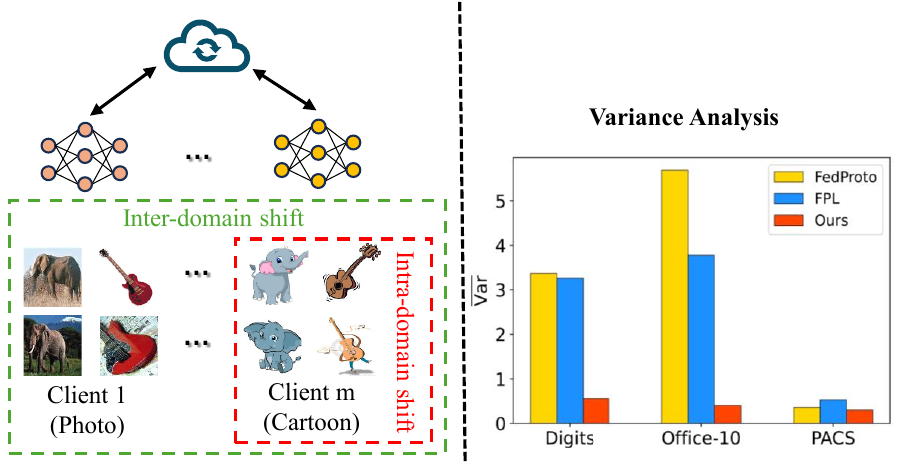}
	\caption{\textbf{Illustration of domain shift challenge in Federated Learning (FL).} We address this from two perspectives: \textcolor{Green}{inter-domain}, referring to distribution shifts across domains, and \textcolor{red}{intra-domain}, capturing variations within the same domain. \textcolor{black}{In the figure (right), we measure the inter-domain prototype variance $\overline{\mathrm{Var}}$, which is defined as the average distance between local prototypes and the generalized prototype for each class, averaged across all clients and classes:  $\mathrm{Var}_k =\frac{1}{M}\sum^M_{m=1}\bigl\lVert p^k_m - g^k\bigr\rVert_2^2$, $\overline{\mathrm{Var}}=\frac{1}{K}\sum_{k=1}^{K}\mathrm{Var}_k$, where $p^k_m$ is the local prototype of class $k$ from client $m$, and $g^k$ is the generalized prototype for class $k$. It can be seen that our method reduces inter-domain prototype variance, highlighting the generalization across domains.}}
	\label{problem_illu} 
\end{figure}

However, most existing FL methods primarily address the label shift, assuming client data is derived from the same domain. In real-world scenarios, private data is often collected from multiple domains. For instance, images of a cat and sketches of a cat might share the same label but come from different domains, leading to heterogeneous feature distributions across clients. Unlike label shift, the impact of domain shift on FL has not been extensively explored.  Under domain shift, a domain gap exists across different participating clients, causing local models to be domain-specific, leading to poor generalization of the global model. To overcome the challenge above, recent FL works on heterogeneous domain~\cite{tan2022federated,huang2023rethinking,wu2024prototype} have considered prototypes, represented as the mean values of vectors within the same semantic class as the solution. These studies~\cite{tan2022federated,tan2022fedproto} obtain the global prototype by averaging local prototypes, which are then used for regularizing the local model to resolve the label shift. Regarding domain shift, the authors in~\cite{huang2023rethinking,wang2024taming} propose a clustering approach to construct unbiased prototypes to provide diverse domain knowledge for multiple clients. 
Clustering methods have demonstrated effectiveness in addressing domain shifts in scenarios where client distributions across domains are non-identical, and it is considered SOTA. However, these methods only consider constructing prototypes at the inter-domain level on the server and overlook the intra-domain perspective of local clients.

Unlike existing federated prototype learning methods, we aim to tackle domain shifts in FL by considering prototypes from intra- and inter-domain perspectives, as shown in Fig.~\ref{problem_illu}. Domain shift in FL can manifest at two levels: inter-domain level, which involves variations between distinct domains (e.g, elephants in Photo and Cartoon domains share same label but have distinct features, as shown in Fig.~\ref{problem_illu}), and intra-domain level, which refers to variation within same domain, such as differences in background, pose. In particular, previous works~\cite{tan2022fedproto,tan2022federated} consider averaging the local prototypes that belong to the same class space to obtain the global prototypes. However, under the domain shift challenge, directly averaging prototypes can create biased global prototypes, similar to the problem in FedAvg with model parameter averaging. A key challenge remains: high variance among inter-domain prototypes, where same-class prototypes from different clients diverge due to domain gaps. Fig.~\ref{problem_illu} illustrates the inter-domain variance comparison. This inconsistency degrades global generalization. To overcome this, we aim to explicitly minimize inter-domain variance during aggregation, generating more consistent and robust generalized prototypes.

Building on the issues identified in prior works, we propose a prototype reweighting scheme to refine \emph{inter-domain prototypes} on the server. We first calculate the initial mean of prototypes from different clients within the same semantic class. However, domain variance may skew this mean toward the dominant domain due to client distribution bias. We assert that prototypes further from the initial mean need more weight than those closer to the mean. Therefore, reweighting scheme assigns more weights to a prototype as its distance from the mean increases. By doing so, we can obtain generalized prototypes that provide inter-domain knowledge for local training, consequently improving performance on challenging domains. It is important to note that our generalized prototype construction maintains privacy through multiple averaging operations~\cite{tan2022fedproto}.
In addition, to address the internal variations at the intra-domain level, we introduce the concept of \emph{intra-domain prototypes} for local clients, enriching the local feature diversity during training. Unlike local prototypes that are sent to the server, we define the intra-domain prototypes as being stored and utilized locally on the client side. Specifically, inspired by the MixUp augmentation~\cite{zhang2018mixup} technique, we create intra-domain prototypes with augmented prototypes for each client. By learning from augmented prototypes, local clients can extract more semantic information from their features, enhancing their generalization capability for subsequent prototype aggregation. To effectively handle domain shift in FL, it is essential to combine prototypes from both intra-domain and inter-domain perspectives. Intra-domain prototypes enhance the diversity within a single domain, improving local learning, while inter-domain prototypes facilitate knowledge transfer across multiple domains, enhancing global generalization.
\begin{figure*}[t]
	\centering
	\includegraphics[width=0.95\linewidth]{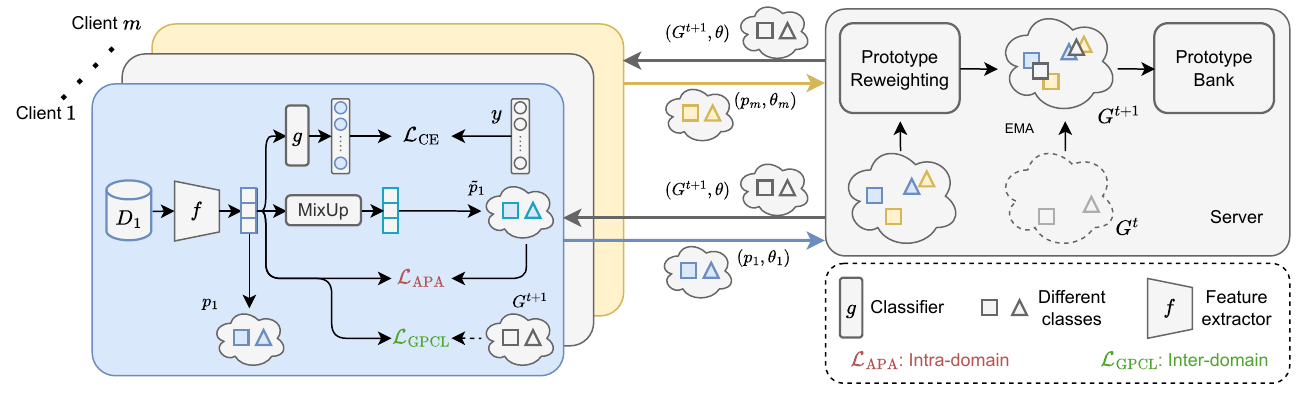}
	\caption{\textbf{Illustration of \textbf{I$^2$PFL}.}  Clients first upload their local prototypes based on Eq.~\ref{eq:localproto} to the server. We introduce the prototype reweighting scheme to generate the Generalized Prototypes $G^{t+1}$ based on Eq.~\ref{eq:gen_proto} and update them with $G^{t}$ from the previous round using the Exponential Moving Average from Eq.~\ref{eq:ema_gen_proto}. We provide inter-domain knowledge from the Generalized Prototypes with $\mathcal{L}_{GPCL}$ from Eq.~\ref{loss_gpc} and enhance the local feature diversity with $\mathcal{L}_{APA}$ based on Eq.~\ref{loss_APA} using the Augmented Prototype from Eq.~\ref{eq:augment_proto}. }
	\label{system} 
\end{figure*}
In this paper, we propose \textbf{I}ntra and \textbf{I}nter-Domain \textbf{P}rototype Federated Learning (I$^2$PFL), which consists of two components:  \textit{Generalized Prototypes Contrastive Learning (GPCL)} and \textit{Augmented Prototype Alignment (APA)}. Our proposed I$^2$PFL is illustrated in Fig.~\ref{system}. 
By combining prototypes at multiple levels, our proposed method enhances the global model's robustness and mitigates negative impacts of domain shift. Our primary contributions are:
\begin{itemize}
\item We focus on FL under the domain shift challenge, recognizing that existing methods primarily address prototype construction at the inter-domain level, overlooking the crucial intra-domain variations within local clients. Our approach uniquely integrates both inter-domain and intra-domain prototypes, offering a more comprehensive solution to domain shift and significantly enhancing the generalization ability of the global model.
\item To tackle the challenge of domain shift in FL, we introduce a novel approach, I$^2$PFL. Our method first introduces prototype learning at the intra-domain level to enhance feature diversity using MixUp-based augmented prototypes. We further construct generalized prototypes with a novel prototype reweighting scheme at the inter-domain level to provide inter-domain knowledge, achieving generalization performance across different domains.
\item We conduct extensive experiments on the Digits, Office-10, and PACS datasets. We demonstrate the superiority of our method over other baselines and validate the effectiveness of each component through ablation studies.
\end{itemize}

\section{Related Work}
\subsection{Federated Learning }
Numerous research efforts in FL~\cite{mcmahan2017communication,li2020federated,tan2022fedproto} have explored diverse techniques to enable the decentralized training of machine learning models while preserving privacy. The conventional FL method FedAvg~\cite{mcmahan2017communication} introduces a training paradigm where the global model is updated by aggregating clients' local model parameters. However, 
FedAvg faces performance degradation when dealing with data heterogeneity. To address the non-iid challenge, some studies incorporate regularization terms to focus on improving the local training, such as FedProx~\cite{li2020federated} with a proximal term calculated by the distance between global and local models and SCAFFOLD~\cite{pmlr-v119-karimireddy20a} with control variates. Other methods, such as FedDyn~\cite{acar2021federated} and pFedMe~\cite{t2020personalized}, also enhance local training through various regularization techniques. Another direction is to improve the aggregation phase. FedMA~\cite{Wang2020Federated} utilizes a Bayesian non-parametric method to average model parameters in a layer-wise manner, while FedAvgM~\cite{hsu2019measuring} incorporates a momentum-based global update at the server. 
However, these methods primarily consider scenarios with single domain data and label skew, overlooking the domain skew challenge in FL. Recently, methods like FedBN~\cite{li2021fedbn} and FPL~\cite{huang2023rethinking} have been developed to address domain skew. Specifically, FPL proposes clustering prototypes to achieve unbiased prototypes, resulting in state-of-the-art performance. Beyond these, several recent works explicitly target domain shift in FL: MPFT fine-tunes server-side adapters with multi-domain prototypes~\cite{zhang2025enhancing}, FDSE reduces client-specific domain skew via domain shift eraser~\cite{wang2025federated} with the personalization-based methods and consensus-based methods. Additionally, COPA~\cite{wu2021collaborative}, FedGA~\cite{zhang2023federated}, FedDG~\cite{liu2021feddg}, and gPerXAN~\cite{le2024efficiently} address the problem of domain generalization, aiming to improve the global model's ability to generalize to unseen domains, i.e., data domains not included in the training process. In contrast, our work tackles a different challenge, focusing on enabling the global model to handle distribution shifts across multiple clients. In this work, we introduce Intra- and Inter-Domain Prototype Federated Learning (I$^2$PFL), which constructs intra- and inter-domain prototypes. Our focus is on enhancing the generalization of the global model under domain shift by utilizing generalized and local augmented prototypes in federated learning.

\subsection{Prototype Learning}
Prototypes~\cite{snell2017prototypical} have achieved success in various applications, including few-shot learning~\cite{tian2020rethinking,zhu2023transductive,zhang2023prototype} and unsupervised learning~\cite{li2021prototypical,gao2023prototype,cui2024effective}. In FL, the concept of prototypes has been extended to address the data heterogeneity challenge~\cite{tan2022federated,qiao2023mp,huang2023rethinking}. FedProto~\cite{tan2022fedproto} was among the first to introduce the use of prototypes in FL, proposing a communication method that exchange prototypes between clients and the server instead of model parameters. 
Recently, FPL~\cite{huang2023rethinking} introduced a cluster-based prototype method to generate unbiased global prototypes, addressing the challenges in FL where the client distributions vary across domains. In addition to FPL, the authors in~\cite{wang2024taming} proposed FedPLVM, which incorporates a dual-level prototype clustering approach and an $\alpha$-sparsity prototype loss to tackle the challenges of learning under domain shift. Concurrently, FedDP proposes domain-independent prototype learning with alignment in both representation and parameter spaces~\cite{fu2025federated}, and MPFT leverages multi-domain prototypes to fine-tune global adapters on the server~\cite{zhang2025enhancing}, providing complementary prototype-centric strategies under domain shift. However, these aforementioned methods primarily focus on constructing the prototypes at the global server, overlooking the intra-domain characteristics of the local clients. In contrast, our approach constructs intra-domain prototypes to increase local feature diversity and introduce a reweighting scheme to inter-domain prototypes, producing the generalized prototypes. The integration of intra- and inter-domain prototypes enables the model to leverage both components effectively: intra-domain prototypes enhance the local generalization within each domain, while inter-domain prototypes provide the shared knowledge across different domains, thus effectively aiding in the generalization of the global model.

\section{Methodology}
\subsection{Overview}
In this paper, we assume there are $M$ clients (indexed by $m$), each with private data $D_m=\{{x}^m_i,{y}^m_i\}$, where $x^m_i$ represents samples and $y^m_i$ denotes the corresponding labels. Under the domain shift, each client has private data with different feature distributions $P_m(x)$, but the label distributions $P_m(y)$ remain the same across multiple clients. 
Client models share the same architecture, consisting of two modules: feature extractor $f$ and classifier $g$. The feature extractor takes the input sample $x^m_i$ and encodes it into a \(d\)-dimensional feature vector  \(h = f(x)\in\mathbb{R}^d\). The classifier $g$ then maps the feature vector $h$ to the logits output $z_{cls}=g(h)\in \mathbb{R}^I$. Given the model parameters for the entire backbone network as $\theta$, and $D=\bigcup_{m=1}^{M} D_m$ representing the sum of samples of all clients, the global objective is formulated, similar to the popular FL framework, FedAvg~\cite{mcmahan2017communication}, as follows: 
\begin{align}
    \argmin_{\theta}{L}(\theta) &= \sum^M_{m=1}\frac{|D_{m}|}{|D|}{\mathcal{L}}_{m}(\theta_{m},D_m),
\end{align}
where the loss function $\mathcal{L}_m$ is the cross-entropy loss $\mathcal{L}_{CE}(z_{cls},y)$ for $m^{th}$ client.
\subsection{Prototype Reweighting Scheme}
Prior research on federated prototype learning~\cite{tan2022fedproto,tan2022federated} typically produce global prototypes by simply averaging the local prototypes from different clients. This can lead to a bias favoring the dominant prototypes and negatively impact performance in domain-skewed FL scenarios. This motivates us to rethink the concept of inter-domain prototypes by designing generalized prototypes that can reduce the bias in prototype averaging and enhance the global model's generalization. We first define the $k^{th}$ class local prototypes from client $m^{th}$ as:
\begin{align}
    p^k_{m}=\frac{1}{|S^k_{m}|}\sum_{i\in S^k_{m}}h_{i},
    \label{eq:localproto}
\end{align}
where $S^k_m$ is the subset of $D_m$ belonging to class $k^{th}$. Then, we further calculate the initial mean of prototypes from different clients within the same semantic class $k^{th}$ as:
\begin{align}
    \begin{split}
    \mu^k&=\frac{1}{M}\sum^M_{m= 1}p^k_m\in\mathbb{R}^d 
    \\
    \mu&=[\mu^1,\mu^2,\dots,\mu^K],
    \end{split}
\end{align}
\begin{figure}[t]
	\centering
	\includegraphics[width=\linewidth]{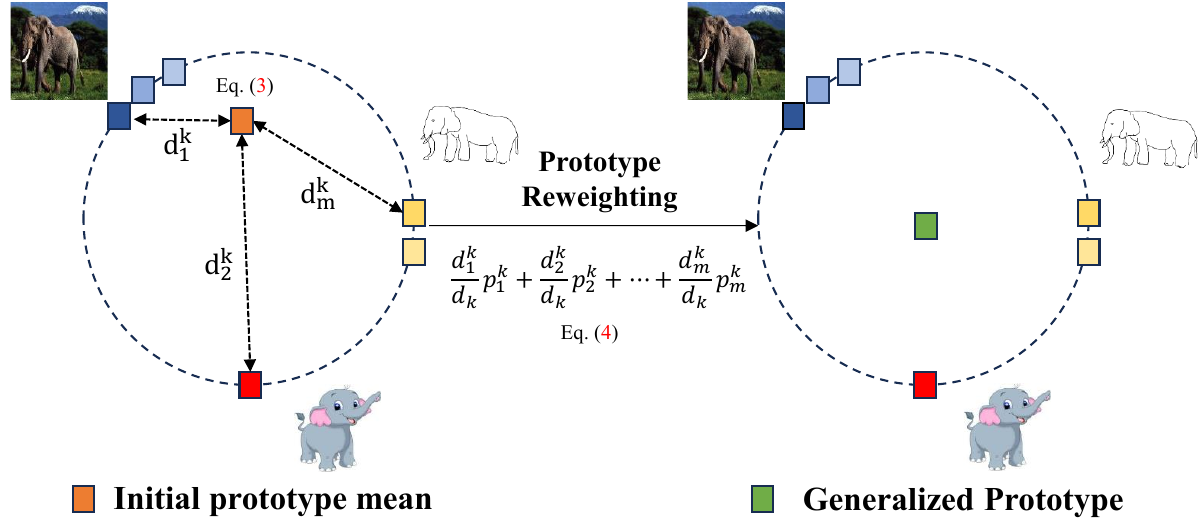}
	\caption{\textbf{Illustration of Prototype Reweighting scheme.} We present the prototype reweighting scheme of the prototypes from different domains in the same semantic class.}
	\label{prototype_reweight} 
\end{figure}

\noindent where $\mu^k$ denotes the initial mean of prototypes belonging to class $k\in K$.

\noindent\textbf{Generalized Prototypes.} Under conditions of domain shift, directly averaging prototypes to obtain an initial mean can lead to significant bias, favoring dominant client prototypes due to discrepancies in client data distributions. We define the distance between the local prototype and the initial mean of prototypes within the same semantic class $k$ as $d^k_m=\|p^k_m-\mu^k\|_2^2$. We assert that prototypes distant from the initial mean indicate important yet underrepresented domain characteristics and thus should be emphasized more in the aggregation process. To achieve a more balanced representation and to reduce domain-specific bias, we propose an adaptive weighting strategy. Specifically, prototypes exhibiting greater distances from the initial mean are assigned higher adaptive weights. This distance-based reweighting ensures that our generalized prototypes represent domain variability comprehensively and robustly, aligning well with variance reduction, as illustrated in Fig.~\ref{prototype_reweight}. We denote the generalized prototypes with our proposed reweighting scheme as follows:
 \begin{align}
    \begin{split}
    g^k&=\sum^M_{m=1}\frac{d^k_{m}}{d^k}p^k_m\in\mathbb{R}^d 
    \\
    G&=[g^1,g^2,\dots,g^K],
    \end{split}
\label{eq:gen_proto}
\end{align}
where $d^k=\sum d_m^k$ denotes the sum of distances between the local prototype and the initial mean of prototypes from different clients, and $g^k$ denotes the generalized prototypes belonging to class $k\in K$. To achieve more stable and consistent generalized prototypes, we apply the Exponential Moving Average (EMA) update to the generalized prototypes of the current communication round $t+1$ from the previous round $t$. The formulation is as follows:

\begin{align}
    G^{t+1}=\beta G^{t+1}+(1-\beta)G^{t},
\label{eq:ema_gen_proto}
\end{align}
where $\beta$ is the decay rate of the EMA update. By applying an EMA update to generalized prototypes and assigning greater weight to past prototypes, we can maintain a balanced representation and mitigate performance fluctuations caused by domain shifts. Compared with the conventional prototype averaging method, our generalized prototypes $G$ achieve fair optimization on multiple domains and avoid bias toward the dominant domain, thus ensuring consistent guidance for the local training process. 

\subsection{Generalized Prototypes Contrastive Learning}
The consistent generalized prototype could enhance the robustness of the global model under the domain shift and guide the local training with inter-domain knowledge. Thus, we apply the contrastive learning between the local features and generalized prototypes. We encourage local features of data samples to closely align with their corresponding generalized prototypes within the same semantic class while pushing away the generalized prototypes of different semantic classes.  Regarding the data samples $\{{x}_i,{y}_i\}$, we first employ the feature extractor to generate the feature vectors 
$h=f(x)\in\mathbb{R}^n$. Let $g$ be the corresponding generalized prototypes $g\in G$, $g^+$ denotes the generalized prototypes with the same semantic class from the local samples. Subsequently, inspired by InfoNCE loss~\cite{oord2018representation}, we design the Generalized Prototype Contrastive Learning (GPCL) loss as follows:
\begin{align}
        \mathcal{L}_{GPCL} = -\frac{1}{B}\sum^B_{i=1}\log\frac{\exp(s(h_{i},g^+)/\tau)}{\sum_{g^k\in G}\exp(s(h_{i},g^k)/\tau)}, 
       \label{loss_gpc}
\end{align}
where $\mathrm{s}( u, v) =  u^\top  v / \lVert u\rVert \lVert v\rVert$ represents the cosine similarity between the local feature and the generalized prototypes, $B$ denotes local batch size and $\tau$ is the temperature parameter. Our target of Eq.~\ref{loss_gpc} is to encourage the local client from different domains to acquire inter-domain knowledge from the generalized prototypes, thereby enhancing the generalization and mitigating the domain shift's negative impact. By doing so, we encourage both generalization and strong class discrimination based on the generalized prototypes.
\subsection{Augmented Prototypes Alignment}

In federated learning, under the domain shift problem, the individual clients possess local data that is limited to a specific domain, which can lead to overfitting and poor generalization. Unlike previous works~\cite{tan2022fedproto,mu2023fedproc,huang2023rethinking} that consider only the prototype construction at the inter-domain level on the server, we propose constructing the intra-domain prototypes at the local clients. To address the limitation of local training data diversity, we conduct the MixUp-based prototype augmentation. Unlike traditional input-level MixUp, which introduces variation in raw input, feature-level MixUp operates on embedding features, producing semantically richer, more stable augmented prototypes that are less domain-specific. We first encode the local samples $\{{x}_i,{y}_i\}$ into the feature vectors $h_i$ using the feature extractor $f$. Inspired by MixUp~\cite{zhang2018mixup} augmentation technique, which generates synthetic instances by combining the features and labels of samples pairs through linear interpolation, we incorporate MixUp strategy to generate the augmented feature as follows:
\begin{align}
    \tilde{h_i}=\gamma h_i+(1-\gamma)h_j,
\end{align}
where $\gamma \sim Beta(\alpha,\alpha)$ with $\alpha \in (0,\infty)$, and $h_j$ is the feature of random data sample $x_j$ from different semantic class on $D_m$. This approach increases the diversity within local features and helps prevent overfitting to data specific to a particular domain. Similar to Eq.~\ref{eq:localproto}, we denote the augmented prototypes of local client $m^{th}$ as:
\begin{align}
\begin{split}
    \tilde{p}^k_{m}&=\frac{1}{|S^k_{m}|}\sum_{i\in S^k_{m}}\tilde{h}_{i}
    \\
    \tilde{p}_m&=[\tilde{p}_m^1,\tilde{p}_m^2,\dots,\tilde{p}_m^K],\end{split}
\label{eq:augment_proto}
\end{align}
\begin{algorithm}[]
    \caption{\textbf{I$^2$PFL}}
    \label{alg:algorithm}
    \textbf{Input}: communication rounds T, local training epochs R, number of clients M, local dataset $D_m$ where $m\in [0, {M-1}]$, feature extractor $f$, classifier $g$.\\
    \textbf{Output}: Global model $\theta_t$
    \begin{algorithmic}[1] 
    \STATE \textbf{Server Execution:} 
    \FOR{$t=0,\dots,{T-1}$}
    \FOR{$m=0,\dots,{M-1}$} 
    \item  $\theta^m_t,p_{m}\leftarrow\textbf{LocalUpdate}(\theta_t,G^t)$
    \ENDFOR
    
    \textcolor{steelblue}{/* Initial mean of prototypes */}
    
    $ \mu^k=\frac{1}{M}\sum_{m\in M}p^k_m\in\mathbb{R}^d$ 

   \textcolor{steelblue}{/* Prototype reweighting */}
   
    $d^k_m=\|p^k_m-\mu^k\|_2^2,~d^k=\sum d_m^k$

    \vspace{0.2em}
    
    $g^k=\sum^M_{m=1}\frac{d^k_{m}}{d^k}p^k_m\in\mathbb{R}^d,~G=[g^1,g^2,\dots,g^K]$ 

    \textcolor{steelblue}{/* EMA update on generalized prototypes */}
    
    $G^{t+1}=\beta G^{t+1}+(1-\beta)G^{t}$

    \textcolor{steelblue}{/* Global model update */}
    
    $\theta_{t+1}\leftarrow\sum_{m=1}^M\frac{|D_m|}{|D|}\theta_t^m$
    \ENDFOR
    \STATE \textbf{Client Execution:}
    \STATE \textbf{LocalUpdate}($\theta_t,G^t$):
    \FOR{$r=0,\dots,R$}
    \FOR{each batch $\in D_m=\{{x}^m_i,{y}^m_i\}$}
    \STATE $h_i=f(x_i),~z_{cls}=g(h_i)$ where $i\in S^k_m$
    \STATE  $\tilde{h_i}=\gamma h_i+(1-\gamma)h_j$ by MixUp augmentation
    \STATE $   \tilde{p}^k_{m}=\frac{1}{|S^k_{m}|}\sum_{i\in S^k_{m}}\tilde{h}_{i},~\tilde{p}_m=[\tilde{p}_m^1,\tilde{p}_m^2,\dots,\tilde{p}_m^K],$
    \STATE $\mathcal{L}_{GPCL}\leftarrow(h_i,G^t)$ in Eq.~\ref{loss_gpc}
    \STATE $\mathcal{L}_{APA}\leftarrow(h_m,\tilde{p}_m)$ in Eq.~\ref{loss_APA}
    \STATE $\mathcal{L}_{CE}\leftarrow(z_{cls},y)$
    \STATE $\mathcal{L}=\mathcal{L}_{CE}+\lambda_{intra}\mathcal{L}_{APA}+\lambda_{inter}\mathcal{L}_{GPCL}$ 
    \STATE $\theta^m_t\leftarrow \theta^m_t-\eta\nabla\mathcal{L}$
    \ENDFOR
    \ENDFOR
    \STATE  $p^k_{m}=\frac{1}{|S^k_{m}|}\sum_{i\in S^k_{m}}h_{i}$
    \STATE $p_{m}=[p^1_{m},p^2_{m},\dots,p^K_{m}]$
    \STATE \textbf{return} $\theta^m_t,p_{m}$      
    \end{algorithmic}
\label{mainalg}
\end{algorithm}

\noindent where $\tilde{p}^k_{m}$ denotes the augmented prototypes of $m^{th}$ client belonging to class $k\in K$. By learning from the augmented prototypes, we enable the model to capture robust representations of intra-domain variations, such as lighting or background differences, thereby improving the generalization capability of the local features and enhancing the robustness of local model training against domain shift. Subsequently, we utilize $\ell_2$ distance and introduce the Augmented Prototype Alignment (APA) as follows:
\begin{align}
    \mathcal{L}_{APA}=\sum_{k}\|h^k_m-\tilde{p}_m^k\|_2^2,
    \label{loss_APA}
\end{align}
where $h^k_m$ is the local features of $k$ semantic class of client $m$. By establishing an alignment between the local representations and the augmented prototypes, we enhance the local feature diversity and avoid overfitting on domain-specific aspects at the intra-domain level. Moreover, it enhances the generalization of the model parameters and prototypes when the global model performs the aggregation on the server. By integrating the prototypes at intra- and inter-domain levels, our proposed scheme enhances the global model's robustness on multiple domains and alleviates the domain shift. We define the overall training objective for each client as follows:
\begin{align}
\mathcal{L}=\mathcal{L}_{CE}+ \underbrace{\lambda_{intra}\mathcal{L}_{APA}}_{\text{\textcolor{red}{\textbf{Intra-domain}}}}+\underbrace{\lambda_{inter}\mathcal{L}_{GPCL}}_{\text{\textcolor{Green}{\textbf{Inter-domain}}}}
 \label{overall_loss}
\end{align}
where $\lambda_{intra}$, $\lambda_{inter}$ are hyper-parameters that control
the importance of $\mathcal{L}_{APA}$, $\mathcal{L}_{GPCL}$, respectively. During the local training phase, each client trains the model on their private data using the loss function specified in Eq.~\ref{overall_loss}. We provide a detailed algorithm of our proposed method in Alg.~\ref{mainalg}. In each communication round, clients receive the generalized prototypes and global model from the server. Then, clients conduct the local training process using augmented and generalized prototypes. After finishing the local training process, the updated local prototypes and local models are sent back to the server, which aggregates them to update the global model and generalized prototypes.

\noindent \textbf{Discussion.} \textcolor{black}{In the previous works, methods such as FPL~\cite{huang2023rethinking} and FedPLVM~\cite{wang2024taming} utilized the clustering method to generate inter-domain prototypes to reduce the bias towards the dominant domain. In contrast, we propose an adaptive distance-based reweighting scheme, which dynamically assigns higher weights to prototypes that are more distant from the initial mean prototype. By adaptively emphasizing these underrepresented prototypes, our approach generates more balanced and generalized inter-domain prototypes, effectively mitigating domain bias arising from domain shift, as demonstrated by the experimental results in Table~\ref{inter_prototypes}. 
Additionally, we incorporate intra-domain prototypes at the local clients to further improve global model generalization under domain shift.} 
\section{Experiment}

\subsection{Experimental Setup}
\noindent \textbf{Datasets.} We conducted experiments using three image classification datasets: ~\textbf{Digits}~\cite{hull1994database,lecun1998gradient,netzer2011reading,roy2018effects}, \textbf{Office-10}~\cite{gong2012geodesic} and \textbf{PACS}~\cite{li2017deeper}. The \textbf{Digits} dataset comprises four domains: MNIST (mt), USPS (up), SVHN (sv), and SYN (syn), each presenting $10$ categories with digit numbers from $0$ to $9$. The \textbf{Office-10} includes four domains: Caltech (C), Amazon (A), Webcam (W), and DSLR (D) of 10 categories. The \textbf{PACS} dataset contains images across $7$ categories from four domains: Photo (P), Art Painting (A), Cartoon (C), and Sketch (S). In the domain shift setting, which is our primary focus, we initialize $20$, $10$, and $10$ clients for Digits, Office-10, and PACS, respectively, and assign domains to clients randomly, following~\cite{huang2023rethinking}. In the domain shift setting, which is our primary focus, we initialize $20$, $10$, and $10$ clients for Digits, Office-10, and PACS, respectively, and assign domains to clients randomly, following~\cite{huang2023rethinking}, as shown in Table~\ref{table:client_distribution}. We sampled a specific proportion from these domains for each client based on task difficulty and dataset size, with sampling rates set at $1\%$, $20\%$, and $30\%$ for Digits, Office-10, and PACS, respectively. To ensure reproducibility, we fixed the seed. Additionally, we evaluate our method in an out-client shift scenario using a leave-one-domain-out evaluation approach. Specifically, we sequentially select one domain as unseen domain while training the model on the remaining domains, treating each domain as a client. The trained model is then evaluated on the unseen domain. 
\begin{table}[]
\centering
\caption{Client distribution for different datasets.}
\resizebox{0.8\linewidth}{!}{
\begin{tabular}{l|cccc}
\toprule
Digits Domains            & \textbf{mt} & \textbf{up} & \textbf{sv} & \textbf{syn} \\ \cmidrule{2-5} 
Client distribution of Digits & 6  & 4  & 3  & 7   \\ \midrule
Office-10 Domains          & \textbf{C}  & \textbf{A}  & \textbf{W}  & \textbf{D}   \\ \cmidrule{2-5} 
Client distribution of Office-10 & 3  & 2  & 1  & 4   \\ \midrule
PACS Domains               & \textbf{P}  & \textbf{A}  & \textbf{C}  & \textbf{S}   \\ \cmidrule{2-5} 
Client distribution of PACS & 3  & 2  & 1  & 4   \\ \bottomrule
\end{tabular}
}
\label{table:client_distribution}
\end{table}
\begin{table*}[]
\begin{center}
\caption{Comparison of our I$^2$PFL against SOTA methods on Digits, Office-10, and PACS datasets under domain shift. Avg denotes the average accuracy ($\%$) across all domains. The best results are marked in~\textbf{bold}.}
\resizebox{18cm}{!}{%
\begin{tabular}{l|ccccc|ccccc|ccccc}
\toprule
\multirow{2}{*}{\textbf{Methods}} & \multicolumn{5}{c|}{Digits} & \multicolumn{5}{c|}{Office-10} & \multicolumn{5}{c}{PACS} \\
                                  & mt  & up  & sv  & \multicolumn{1}{c|}{syn} & \multicolumn{1}{l|}{Avg} & C   & A   & W   & \multicolumn{1}{c|}{D} & \multicolumn{1}{l|}{Avg} & P   & A   & C   & \multicolumn{1}{c|}{S} & \multicolumn{1}{l}{Avg} \\ \midrule
FedAvg~\cite{mcmahan2017communication}        & 97.85  & 90.76  & 80.52  & \multicolumn{1}{c|}{\underline{73.30}}  & 85.61  & 64.91  & 76.32  & 42.76  & \multicolumn{1}{c|}{46.00}  & 57.50  & 81.65  & 68.07  & 72.84  & \multicolumn{1}{c|}{87.14}  & 77.43  \\
FedProx~\cite{li2020federated}                & 98.10  & 90.76  & 81.26  & \multicolumn{1}{c|}{73.05}  & 85.79  & 64.55  & 77.16  & 54.14  & \multicolumn{1}{c|}{45.33}  & 60.30  & 80.67  & 67.59  & 75.41  & \multicolumn{1}{c|}{88.92}  & 78.15  \\
FedDyn~\cite{acar2021federated}                & 98.16  & 90.72  & 81.30  & \multicolumn{1}{c|}{72.36}  & 85.64  & 63.57  & 76.95  & 55.52  & \multicolumn{1}{c|}{42.00}  & 59.51  & 83.27  & 67.85  & 74.44  & \multicolumn{1}{c|}{88.36}  & 78.48  \\
MOON~\cite{li2021model}                      & 97.77  & \underline{91.80}  & 82.22  & \multicolumn{1}{c|}{60.77}  & 83.14  & 61.61  & 74.11  & 48.97  & \multicolumn{1}{c|}{46.67}  & 57.84  & 84.64  & \underline{73.21}  & 74.70  & \multicolumn{1}{c|}{91.85}  & 81.10  \\
FedProc~\cite{mu2023fedproc}             & 97.83  & 90.28  & 81.09  & \multicolumn{1}{c|}{68.10}  & 84.33  & 62.23  & 78.00  & 44.83  & \multicolumn{1}{c|}{33.33}  & 54.62  & 83.18  & 70.27  & 75.23  & \multicolumn{1}{c|}{\textbf{94.29}}  & 80.71  \\
FedProto~\cite{tan2022fedproto}                & 98.10  & 91.48  & 81.70  & \multicolumn{1}{c|}{72.95}  & 86.05  & 65.89  & \underline{79.16}  & 58.27  & \multicolumn{1}{c|}{56.65}  & 64.99  & \textbf{89.29}  & 71.08  & 73.59  & \multicolumn{1}{c|}{87.83}  & 80.45  \\
FPL~\cite{huang2023rethinking}                & 98.18  & 91.24  & \underline{82.37}  & \multicolumn{1}{c|}{72.97}  & 86.19  & \underline{69.02}  & 79.05  & \underline{65.52}  & \multicolumn{1}{c|}{53.33}  & 66.73  & 85.27  & 71.40  & 74.96  & \multicolumn{1}{c|}{90.83}  & 80.62  \\ 
FedPLVM~\cite{wang2024taming} &  \underline{98.26}     & 90.98      & 82.00      &  \multicolumn{1}{c|}{\textbf{74.19}}       &  \underline{86.36}     &68.93      & 78.74      &  62.41     & \multicolumn{1}{c|}{\underline{61.33}}       & \underline{67.85}      &  86.70     & 73.00      & \textbf{76.86}      & \multicolumn{1}{c|}{90.64}       &   \underline{81.80}    \\ \midrule
\textbf{I$^2$PFL}       & \textbf{98.32} & \textbf{93.33} & \textbf{84.65} & \multicolumn{1}{c|}{73.02} & \textbf{87.33} & \textbf{71.52}  & \textbf{81.79}  & \textbf{72.07}  & \multicolumn{1}{c|}{\textbf{66.00}}  & \textbf{72.84}  & \underline{87.85}  & \textbf{73.29}  & \underline{75.66}  & \multicolumn{1}{c|}{\underline{92.20}}  & \textbf{82.25}  \\ 
\bottomrule
\end{tabular}
}%
\label{comparison_main}
\end{center}
\end{table*}

\begin{table*}[]
\begin{center}
\caption{Comparison of our I$^2$PFL against SOTA methods on Digits, Office-10, and PACS datasets under out-client shift setting. Avg denotes the average accuracy ($\%$) across different unseen domains. The best results are marked in~\textbf{bold}.}
\resizebox{18cm}{!}{%
\begin{tabular}{l|l|ccccc|ccccc|ccccc}
\toprule
\multicolumn{2}{c|}{\textbf{Methods}} & \multicolumn{5}{c|}{Digits} & \multicolumn{5}{c|}{Office-10} & \multicolumn{5}{c}{PACS} \\
\multicolumn{2}{c|}{} & $\rightarrow$ mt  & $\rightarrow$ up  & $\rightarrow$ sv  & \multicolumn{1}{c|}{$\rightarrow$ syn} & \multicolumn{1}{l|}{Avg} & $\rightarrow$ C  & $\rightarrow$ A  & $\rightarrow$ W  & \multicolumn{1}{c|}{$\rightarrow$ D} & \multicolumn{1}{l|}{Avg} & $\rightarrow$ P   & $\rightarrow$ A   & $\rightarrow$ C   & \multicolumn{1}{c|}{$\rightarrow$ S} & \multicolumn{1}{l}{Avg} \\ \midrule
\multirow{4}{*}{FL} & FedAvg~\cite{mcmahan2017communication}        &72.02  &82.90   &64.27   &\multicolumn{1}{c|}{75.89}  &73.77   & 42.58  & 66.62  & 61.42  & \multicolumn{1}{c|}{70.43} & 60.26   & 72.34  & 67.15  & 65.63  & \multicolumn{1}{c|}{73.52}  & 69.66  \\ 
                    & FedProx~\cite{li2020federated}       & 70.36  & 83.30  & 64.56  & \multicolumn{1}{c|}{76.98}  & 73.80  & 50.65  & 64.42  & 61.56  & \multicolumn{1}{c|}{69.96}  & 61.65 & 69.22  & 68.70  & 66.36  & \multicolumn{1}{c|}{74.48}  & 69.69  \\ 
                    & FedDyn~\cite{acar2021federated}        &73.16  & 82.25  & 65.56  & \multicolumn{1}{c|}{78.79}  & 74.94  & 49.33  & 67.70  & 59.48  & \multicolumn{1}{c|}{70.89}  & 61.85  & 72.47  & 68.96  & 66.78  & \multicolumn{1}{c|}{73.46}  & 70.42 \\ 
                    & MOON~\cite{li2021model}          & 68.72  & 76.26  & 62.91  & \multicolumn{1}{c|}{71.89}  & 69.95  & 45.64  & 60.43  & 58.67  & \multicolumn{1}{c|}{69.17}  & 58.48  & 69.71  & 64.67  & 66.09  & \multicolumn{1}{c|}{72.20}  & 68.17  \\ \midrule
\multirow{2}{*}{FL + DG} & COPA~\cite{wu2021collaborative}        & 73.00  &82.95  &62.33   & \multicolumn{1}{c|}{\textbf{83.25}}  & 75.38   & 50.53  & 67.95  & 62.84  & \multicolumn{1}{c|}{70.14}  & 62.87  & 71.16  & 65.24  & 71.62  & \multicolumn{1}{c|}{75.54}  & 70.19  \\  
                    & FedGA~\cite{zhang2023federated}       & 73.85   & 83.24  &67.30   & \multicolumn{1}{c|}{80.37}  & 76.19   & 48.26  & 65.83  & 63.32  & \multicolumn{1}{c|}{67.51}  & 61.23  & 71.19  & 66.53  &  \underline{72.25}  & \multicolumn{1}{c|}{74.19}  & 71.04  \\  \midrule
\multirow{5}{*}{Prototype-based FL} & FedProc~\cite{mu2023fedproc}   & 64.51  & 78.89  & 52.86  & \multicolumn{1}{c|}{80.63}  & 69.22  & 46.40  & 59.25  & 55.64  & \multicolumn{1}{c|}{69.67}  & 57.74  & 72.24  & \underline{72.27}  & 68.71  & \multicolumn{1}{c|}{\textbf{76.54}}  & 72.44  \\  
                    & FedProto~\cite{tan2022fedproto}    & 73.72  & 82.42  & 67.90  & \multicolumn{1}{c|}{77.59}  & 75.41  & \underline{51.25}  & 69.33  & 64.48  & \multicolumn{1}{c|}{71.18}  & 64.06  & 71.47  & 69.33  & 70.33  & \multicolumn{1}{c|}{73.96}  & 71.27  \\  
                    & FPL~\cite{huang2023rethinking}         & 73.87  & 83.72  & \textbf{70.21}  & \multicolumn{1}{c|}{79.56}  & 76.84  & 43.88  & \underline{71.19}  & 62.12  & \multicolumn{1}{c|}{\underline{73.13}}  & 62.58  & 73.83  & 68.48  & 71.26  & \multicolumn{1}{c|}{\underline{75.58}}  & 72.29  \\  
                    & FedPLVM~\cite{wang2024taming}     & \underline{76.31}  & \underline{84.05}  & 66.40  & \multicolumn{1}{c|}{81.73}  & \underline{77.12}  & 50.85  & 70.66  & \textbf{66.22}  & \multicolumn{1}{c|}{73.00}  & \underline{65.18}  & \underline{77.14}  & \textbf{74.26}  & 65.08  & \multicolumn{1}{c|}{74.87}  & \underline{72.84}  \\  
                    & \textbf{I$^2$PFL}  & \textbf{77.25} & \textbf{84.16} & \underline{69.36} & \multicolumn{1}{c|}{\underline{81.86}} & \textbf{78.16}  & \textbf{51.47} & \textbf{71.62} & \underline{65.96} & \multicolumn{1}{c|}{\textbf{73.70}} & \textbf{65.69}  & \textbf{78.81}  & 70.53  & \textbf{72.74}  & \multicolumn{1}{c|}{74.99}  & \textbf{74.27}  \\ 
\bottomrule
\end{tabular}
}%
\label{comparison_digits_office_pacs_unseen}
\end{center}
\end{table*}



\noindent \textbf{Model Architecture.} 
For the Digits and Office-10 datasets, we used ResNet-10~\cite{he2016deep} as the base model architecture, while for the PACS dataset, we employed ResNet-18~\cite{he2016deep}.

\noindent \textbf{Implementation Details.} The communication round is set to $100$, and the local training epoch is $10$ for all datasets. We employ the SGD~\cite{robbins1951stochastic} optimizer with a weight decay of $1e-5$ and a learning rate of $0.01$ across all datasets. The training batch size is $32$ for the Digits and Office-10 datasets, and $16$ for the PACS dataset. The EMA $\beta$ is set as 0.99 for all datasets. Top-1 accuracy is used as the evaluation metric. Each experiment is repeated three times, and we report the mean values from the last $5$ communication rounds.

\begin{figure}[]
    \centering 
    \begin{subfigure}[b]{0.93\linewidth} 
        \centering
        \begin{minipage}{0.35\linewidth}
            \includegraphics[width=\linewidth]{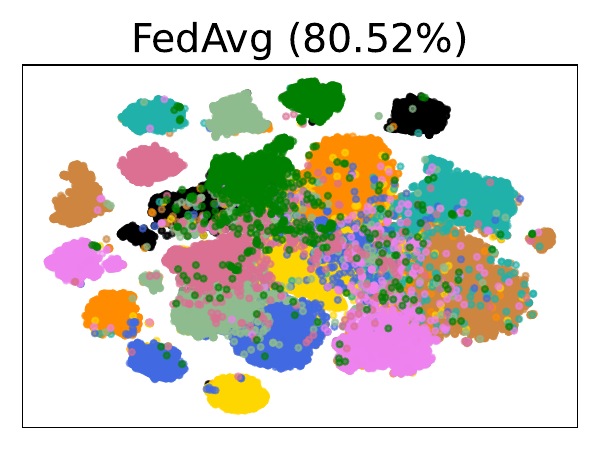}
        \end{minipage}%
        \begin{minipage}{0.35\linewidth}
            \includegraphics[width=\linewidth]{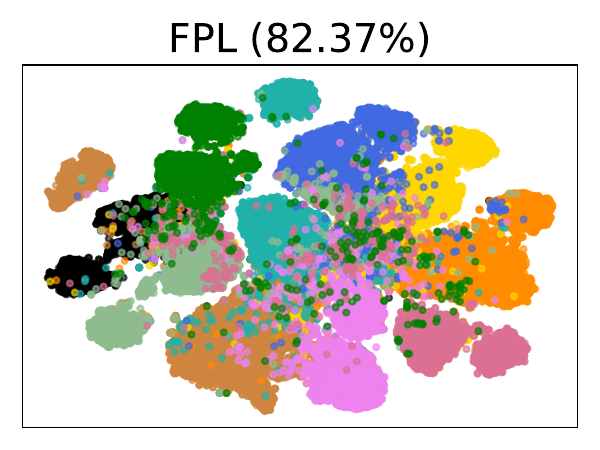}
        \end{minipage}%
        \begin{minipage}{0.35\linewidth}
            \includegraphics[width=\linewidth]{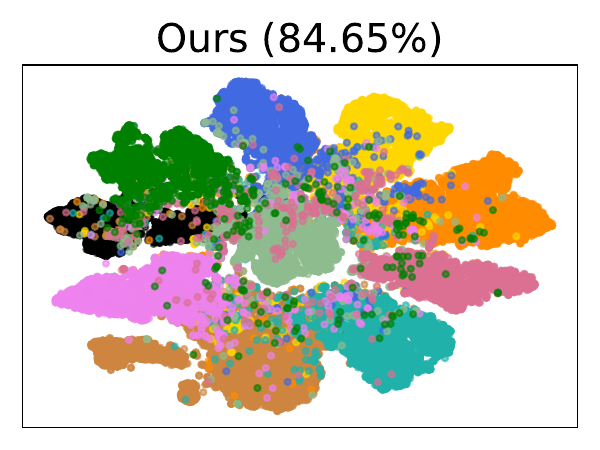}
        \end{minipage}
        \caption{SVHN}
    \end{subfigure}
    
    \begin{subfigure}[b]{0.93\linewidth}
        \centering
        \begin{minipage}{0.35\linewidth}
            \includegraphics[width=\linewidth]{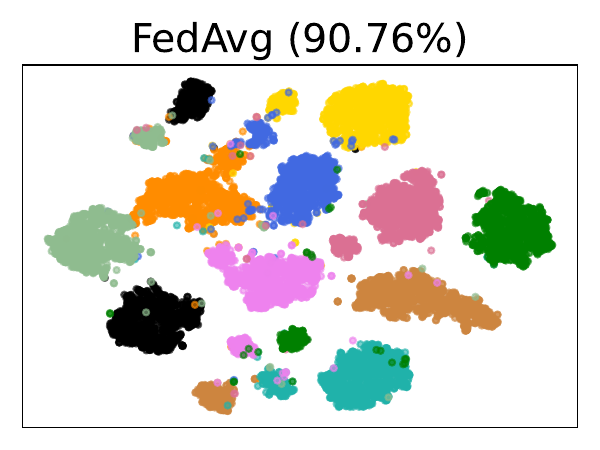}
        \end{minipage}%
        \begin{minipage}{0.35\linewidth}
            \includegraphics[width=\linewidth]{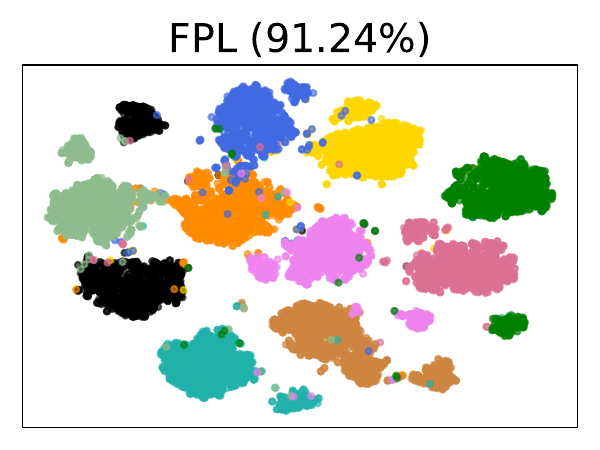}
        \end{minipage}%
        \begin{minipage}{0.35\linewidth}
            \includegraphics[width=\linewidth]{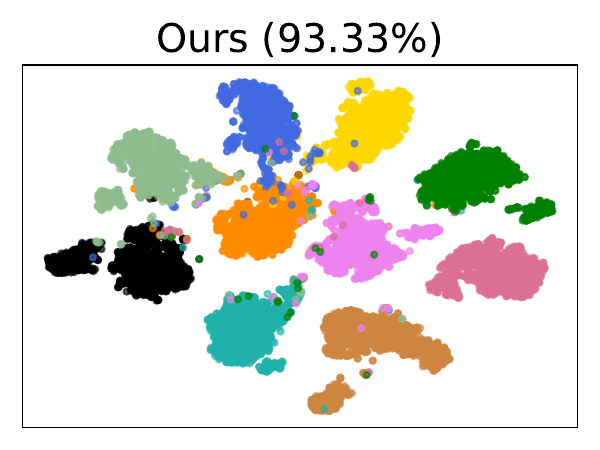}
        \end{minipage}
        \caption{USPS}
    \end{subfigure}

    \caption{t-SNE Visualization of features in the Digits dataset.}
    \label{fig:visualization}
\end{figure}

\noindent \textbf{Baselines.} For evaluation, we compare our \textbf{I$^2$PFL} against several state-of-the art FL methods: \textbf{FedAvg}~(AISTATS'17)~\cite{mcmahan2017communication}, \textbf{FedProx}~(MLsys'21)~\cite{li2020federated}, \textbf{FedDyn}~(ICLR'21)~\cite{acar2021federated}, \textbf{MOON}~(CVPR'21)~\cite{li2021model}, as well as prototype-based FL methods: \textbf{FedProc}~(FGCS'23)~\cite{mu2023fedproc}, \textbf{FedProto}~(AAAI'22)~\cite{tan2022fedproto} (with parameter averaging), ~\textbf{FPL}~(CVPR'23)~\cite{huang2023rethinking}, and~\textbf{FedPLVM}~(NeurIPS'24)~\cite{wang2024taming}. For the out-client shift setting, we include SOTA baselines from Federated Domain Generalization setting, such as \textbf{COPA}~(ICCV'21)~\cite{wu2021collaborative} and \textbf{FedGA}~(CVPR'23)~\cite{zhang2023federated}.

\begin{figure*}[]
    \centering
    \begin{subfigure}{0.33\textwidth}
        \centering
        \includegraphics[width=\linewidth]{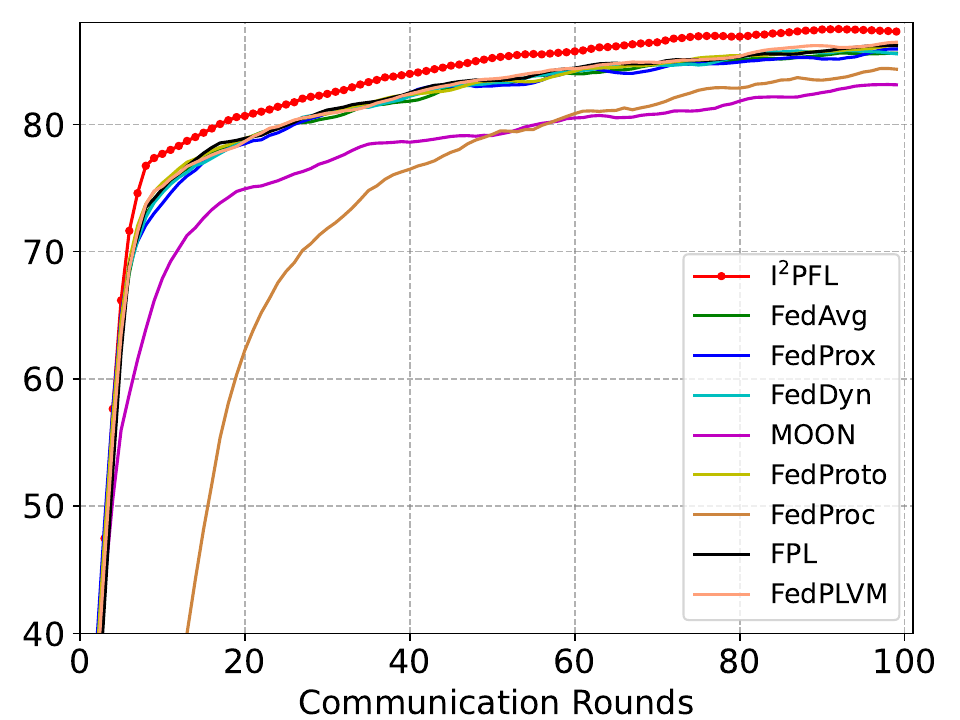}
        \caption{Digits}
    \end{subfigure}%
    \hfill
    \begin{subfigure}{0.33\textwidth}
        \centering
        \includegraphics[width=\linewidth]{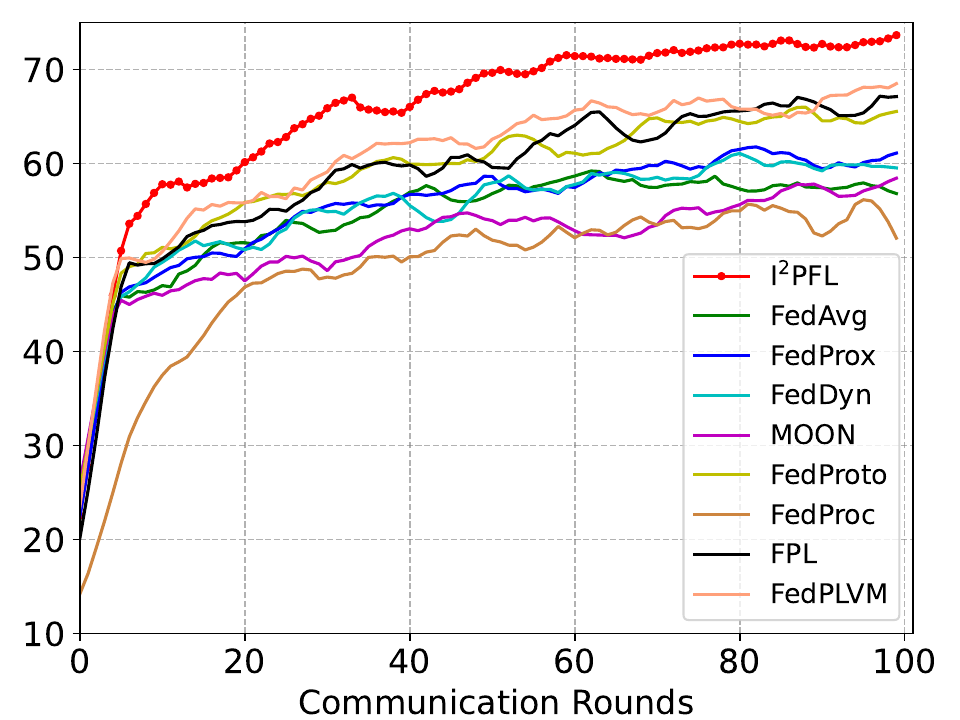}
        \caption{Office-10}
    \end{subfigure}%
    \hfill
    \begin{subfigure}{0.33\textwidth}
        \centering
        \includegraphics[width=\linewidth]{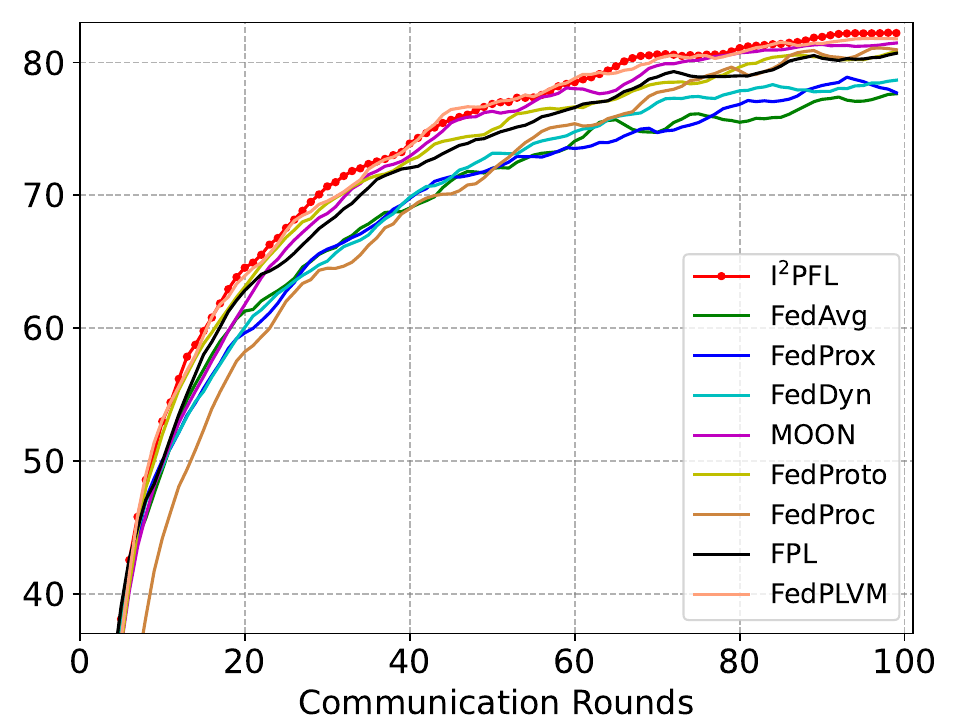}
        \caption{PACS}
    \end{subfigure}
    \caption{Visualization of training curves of average test accuracy on three datasets under the domain shift setting.}
    \label{curve}
\end{figure*}

\subsection{Performance Comparison}
\noindent \textbf{Comparison to SOTA methods.}
Table~\ref{comparison_main} presents the performance comparison of our proposed I$^2$PFL with other SOTA methods on three datasets. As the results show, I$^2$PFL consistently outperforms other baselines across multiple domains. The average accuracy depicts the effectiveness in achieving better generalization. For the Digits dataset, I$^2$PFL demonstrates superior performance across all domains, with an average accuracy improvement of $0.97\%$ compared to the second best method FedPLVM. Regarding the Office-10 dataset, our method outperforms the state-of-the-art methods FPL and FedPLVM by a notable gap, illustrating an improvement of $4.99\%$. Specifically, we improve the performance on a challenging domain like DSLR, where I$^2$PFL significantly outperforms other methods. In the PACS dataset, methods incorporating contrastive learning tend to achieve higher average accuracy across all domains. However, I$^2$PFL still outperforms other approaches in most domains, with a $1.15\%$ improvement in average accuracy compared to MOON. By integrating intra- and inter-domain prototypes, we enhance the generalization across multiple domains, effectively avoiding the bias toward any specific domain. Regarding the out-client shift setting, as shown in Table~\ref{comparison_digits_office_pacs_unseen}, our proposed method I$^2$PFL achieves average accuracy across unseen clients of $78.16\%$ on the Digits dataset, $65.69\%$ on the Office-10 dataset and $74.27\%$ on the PACS dataset, surpassing the second-best methods by $1.04\%$, $0.51\%$ and $1.43\%$, respectively. These results demonstrate the strong generalization capability of our method to unseen domains, outperforming other state-of-the-art techniques. Our approach, I$^2$PFL, incorporates both intra-domain and inter-domain prototypes, significantly improving the model’s generalization ability to the unseen domain during training. 

We illustrate the representations produced by our I$^2$PFL using t-SNE~\cite{van2008visualizing} on Digits datasets, as shown in Fig.~\ref{fig:visualization}. We compare the representations extracted from the global model between our proposed method, FPL and FedAvg on Digits dataset with SVHN and USPS domains. The figures show that the features generated by our method are more distinctly separated compared to those from other methods, illustrating the better generalization of the global model across different domains.

\noindent \textbf{Convergence Analysis.} 
Fig.~\ref{curve} depicts the performance curves of our methods and SOTA baselines on all datasets under the domain shift setting. We clearly observe that I$^2$PFL not only converges faster but also exhibits significantly more stable training behavior and reduced fluctuations compared to other methods. This empirical evidence highlights the robustness of our proposed method, demonstrating their effectiveness in stabilizing model training and convergence in the presence of domain shift.


\begin{table}[]
\centering
\caption{Ablation study on key components of our I$^2$PFL across datasets.}
\resizebox{8cm}{!}{%
\begin{tabular}{c|ccccc}
\toprule
\multirow{2}{*}{\textbf{Methods}}       & \multicolumn{5}{c}{Digits}                                                                                                                 \\
                               & mt                        & up                        & sv                        & \multicolumn{1}{c|}{syn}   & Avg                       \\ \midrule
\texttt{w/o} ($\mathcal{L}_{GPCL}$, $\mathcal{L}_{APA}$)             & 97.85                     & 90.76                     & 80.52                     & \multicolumn{1}{c|}{73.30} & 85.61                     \\
 \texttt{w/o} $\mathcal{L}_{APA}$              & 98.17                     & 91.07                     & 82.45                     & \multicolumn{1}{c|}{72.75} & 86.11                     \\
\texttt{w/o} $\mathcal{L}_{GPCL}$             & 98.15                     & 92.77                     & 83.15                     & \multicolumn{1}{c|}{73.12} & 86.80                     \\
\texttt{w/o} EMA                &  97.93                    & 92.03                     &  82.27                    & \multicolumn{1}{c|}{72.74} &  86.24           \\  
Ours                        & 98.32                          & 93.33                          &84.65                           & \multicolumn{1}{c|}{73.02}      &    \textbf{87.33}                       \\ \midrule
\multirow{2}{*}{\textbf{Methods}}       & \multicolumn{5}{c}{Office-10}                                                                                                              \\
                               & C                         & A                         & W                         & \multicolumn{1}{c|}{D}    & Avg                       \\ \midrule
\texttt{w/o} ($\mathcal{L}_{GPCL}$, $\mathcal{L}_{APA}$)            & 64.91                     & 76.32                     & 42.76                     & \multicolumn{1}{c|}{46.00} & 57.50                     \\
 \texttt{w/o} $\mathcal{L}_{APA}$              & 68.03                     & 78.10                     & 44.48                     & \multicolumn{1}{c|}{53.99} & 61.15                     \\
\texttt{w/o} $\mathcal{L}_{GPCL}$               & 63.75                     & 79.16                     & 55.52                     & \multicolumn{1}{c|}{58.00} & 64.11                     \\
\texttt{w/o} EMA                 &   66.16                   &  80.79                   & 70.45                   & \multicolumn{1}{c|}{68.00} & 71.35           \\  
Ours                        & 71.52                          & 81.79                           &  72.07                           & \multicolumn{1}{c|}{66.00}      &  \textbf{72.84}                          \\ \midrule
\multirow{2}{*}{\textbf{Methods}}       & \multicolumn{5}{c}{PACS}                                                                                                                   \\
                               & P                         & A                         & C                         & \multicolumn{1}{c|}{S}    & Avg                       \\ \midrule
\texttt{w/o} ($\mathcal{L}_{GPCL}$, $\mathcal{L}_{APA}$)             & 81.65                     & 68.07                     & 72.84                     & \multicolumn{1}{c|}{87.14} & 77.43                     \\
 \texttt{w/o} $\mathcal{L}_{APA}$               & 85.43                     & 68.49                     & 75.25                     & \multicolumn{1}{c|}{88.31} & 79.37                     \\
\texttt{w/o} $\mathcal{L}_{GPCL}$           & 85.33                     & 71.64                     & 75.89                     & \multicolumn{1}{c|}{89.87} & 80.68                     \\
\texttt{w/o} EMA                & 87.00                     & 69.64                     & 73.79                    & \multicolumn{1}{c|}{87.53} & 79.49           \\  
Ours                       & 87.85                          &  73.29                         & 75.66                           & \multicolumn{1}{c|}{92.20}      & \textbf{82.25}                           \\ \bottomrule
\end{tabular}
}%
\label{ablation_component}
\end{table}

\subsection{Ablation Study and Analysis}\label{sec:ablation}
\noindent \textbf{Contributions of Key Components.} To evaluate the effect of each component on I$^2$PFL's performance, we perform an ablation study by selectively removing individual components, as detailed in Table~\ref{ablation_component}. Our results show that both GPCL and APA improve performance over the baseline, highlighting the value of intra-domain and inter-domain prototypes.  Notably, APA significantly impacts performance across all datasets, demonstrating the effectiveness of enhancing the feature diversity of our proposed intra-domain prototypes on the local side. Additionally, we evaluate the impact of using EMA updates for generalized prototypes, which improves performance across all datasets by smoothing the prototype updates over time and reducing fluctuations caused by varying domain distributions. These observations highlight the critical importance of leveraging both intra- and inter-domain prototypes to improve the generalization of the global model under domain shift.

\begin{table}[]
\centering
\caption{Performance analysis of I$^2$PFL with different inter-domain prototype construction methods.}
\resizebox{8.5cm}{!}{%
\begin{tabular}{l|cc|cc|cc}
\toprule
\multirow{2}{*}{\textbf{Inter-domain Prototypes}} 
  & \multicolumn{2}{c|}{Digits} 
  & \multicolumn{2}{c|}{Office-10} 
  & \multicolumn{2}{c}{PACS} \\
  & Avg & $\Delta$ & Avg & $\Delta$ & Avg & $\Delta$ \\
\midrule
Averaging           & 86.37 & --    & 69.68 & --    & 79.90 & --    \\
Clustering          & 86.52 & +0.15 & 71.84 & +2.16 & 80.16 & +0.26 \\
Reweighting (Ours)  & \textbf{87.33} & \textbf{+0.96} 
                    & \textbf{72.84} & \textbf{+3.16} 
                    & \textbf{82.25} & \textbf{+2.35} \\
\bottomrule
\end{tabular}%
}
\label{inter_prototypes}
\end{table}

\begin{figure*}[]
    \centering
    \begin{subfigure}{0.32\textwidth}
        \centering
        \includegraphics[width=\linewidth]{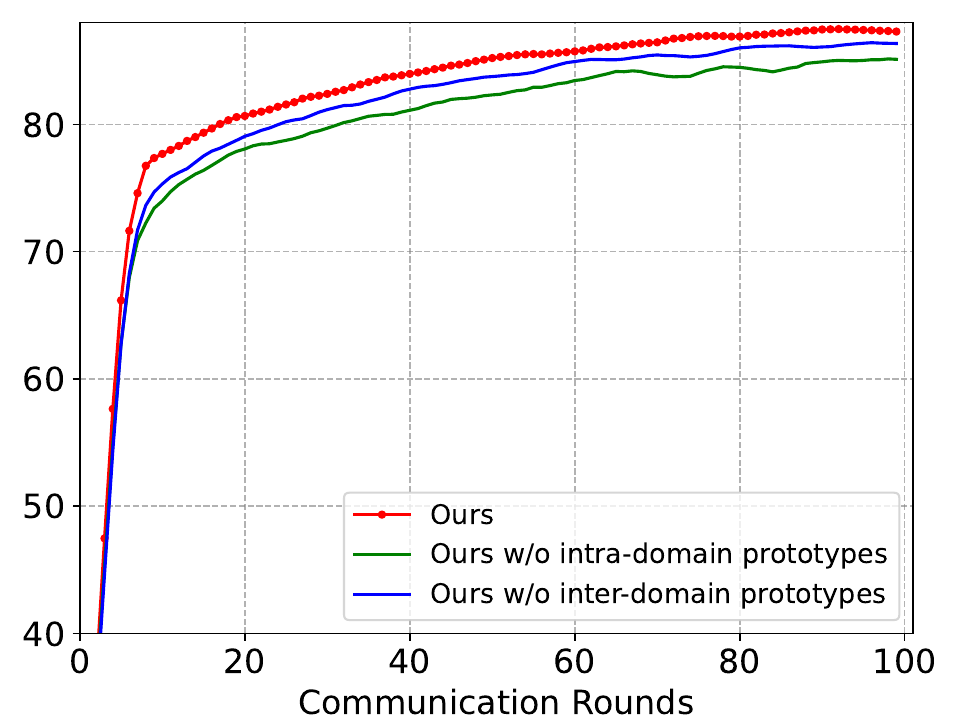}
        \caption{Digits}
        \label{fig:temperature_ucihar}
    \end{subfigure}%
    \hfill
    \begin{subfigure}{0.32\textwidth}
        \centering
        \includegraphics[width=\linewidth]{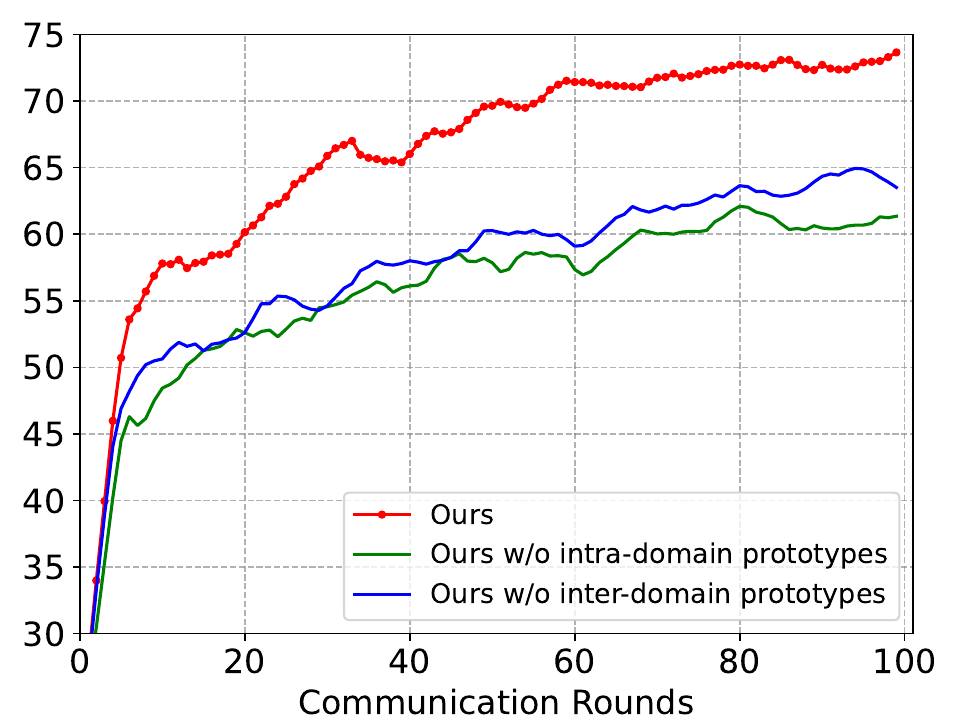}
        \caption{Office-10}
        \label{fig:dimension_hatefulmemes}
    \end{subfigure}%
    \hfill
    \begin{subfigure}{0.32\textwidth}
        \centering
        \includegraphics[width=\linewidth]{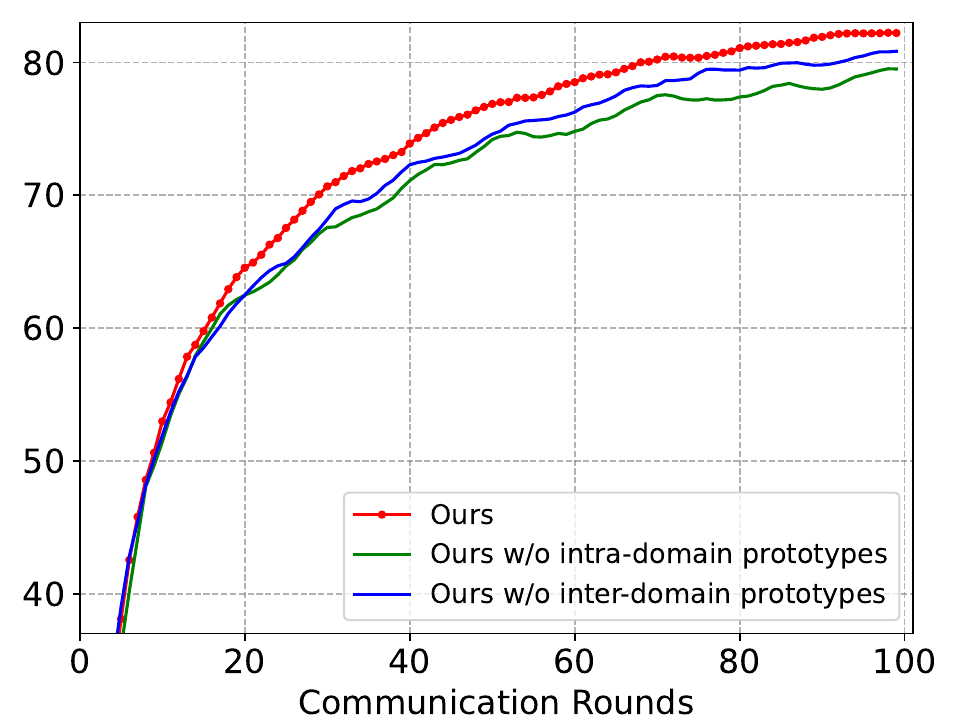}
        \caption{PACS}
        \label{fig:dimension_meld}
    \end{subfigure}
    \caption{Effect of different prototype components across three datasets.}
    \label{curve_prototypes}
\end{figure*}

\begin{table}[]
\centering
\caption{Performance of differential privacy (DP) with noise parameters across Digits, Office-10, and PACS datasets.}
\resizebox{8cm}{!}{%
\begin{tabular}{l|ccccc}
\toprule
\multicolumn{1}{c|}{\multirow{2}{*}{\textbf{Local Prototypes}}} & \multicolumn{5}{c}{Digits} \\ \multicolumn{1}{c|}{}  
                                           & mt & up & sv & \multicolumn{1}{c|}{syn} & Avg \\
\midrule
\texttt{w} DP       & 98.14 & 93.01 & \textbf{84.67} & \multicolumn{1}{c|}{72.78} & 87.15 \\
\texttt{w/o} DP     & \textbf{98.32} & \textbf{93.33} & 84.65 & \multicolumn{1}{c|}{\textbf{73.02}}& \textbf{87.33} \\
\midrule
\multicolumn{1}{c|}{\multirow{2}{*}{\textbf{Local Prototypes}}} & \multicolumn{5}{c}{Office-10} \\ \multicolumn{1}{c|}{} 
                                           & C & A & W & \multicolumn{1}{c|}{D} & Avg \\
\midrule
\texttt{w} DP       & 68.66 & \textbf{82.53} & 68.90 & \multicolumn{1}{c|}{62.00} & 72.02 \\
\texttt{w/o} DP     & \textbf{71.52} & 81.79 & \textbf{72.07} & \multicolumn{1}{c|}{\textbf{66.00}} & \textbf{72.84} \\
\midrule
\multicolumn{1}{c|}{\multirow{2}{*}{\textbf{Local Prototypes}}} & \multicolumn{5}{c}{PACS} \\ \multicolumn{1}{c|}{} 
                                           & P & A & C & \multicolumn{1}{c|}{S} & Avg \\
\midrule
\texttt{w} DP       & \textbf{89.18} & 70.69 & 75.55 & \multicolumn{1}{c|}{88.05} & 80.87 \\
\texttt{w/o} DP     & 87.85 & \textbf{73.29} & \textbf{75.66} & \multicolumn{1}{c|}{\textbf{92.20}} & \textbf{82.25} \\
\bottomrule
\end{tabular}
}
\label{dp_performance}
\end{table}

\begin{table}[]
\centering
\caption{Ablation study on the effect of MixUp on intra-domain prototypes in the Office-10 and PACS datasets.}
\resizebox{8cm}{!}{%
\textcolor{black}{
\begin{tabular}{l|ccccc}
\toprule
\multicolumn{1}{c|}{\multirow{2}{*}{\begin{tabular}[c]{@{}c@{}}\textbf{Intra-domain} \\ \textbf{Prototypes}\end{tabular}}} & \multicolumn{5}{c}{Office-10}                                       \\
\multicolumn{1}{c|}{}                                                                                    & C     & A     & W     & \multicolumn{1}{c|}{D}     & Avg            \\ \midrule
\texttt{w/o} MixUp                                                                                                & 68.57 & 79.16 & 68.28 & \multicolumn{1}{c|}{48.00} & 66.00          \\
MixUp (Input)                                                                                      & 66.62 & 79.52 & 65.43 & \multicolumn{1}{c|}{62.67} & 68.56          \\
Ours                                                                                                     & 71.52 & 81.79 & 72.07 & \multicolumn{1}{c|}{66.00} & \textbf{72.84} \\ \midrule
\multicolumn{1}{c|}{\multirow{2}{*}{\begin{tabular}[c]{@{}c@{}}\textbf{Intra-domain} \\ \textbf{Prototypes}\end{tabular}}} & \multicolumn{5}{c}{PACS}                                             \\
\multicolumn{1}{c|}{}                                                                                    & P     & A     & C     & \multicolumn{1}{c|}{S}     & Avg            \\ \midrule
\texttt{w/o} MixUp                                                                                                & 85.41        & 70.53      & 71.85      & \multicolumn{1}{c|}{91.57}      &  79.84              \\
MixUp (Input)                                                                                            & 87.13      &  71.67     & 73.22      & \multicolumn{1}{c|}{91.82}      &  80.96              \\
Ours                                                                                                     &  87.85     & 73.29      & 75.66      &  \multicolumn{1}{c|}{92.20}      & \textbf{82.25}               \\ \bottomrule
\end{tabular}}
}%
\label{tab:intra_domain_comparison}
\end{table}

\noindent \textbf{Analysis on inter-domain prototypes.}
\textcolor{black}{In Table~\ref{inter_prototypes}, we evaluate the effectiveness of our proposed inter-domain prototypes with prototype reweighting scheme against the prototype averaging method and the FINCH~\cite{sarfraz2019efficient} clustering method used in FPL~\cite{huang2023rethinking}}. The results clearly demonstrate the superior performance of utilizing our reweighting approach, showing improvements of $0.96\%$, $3.16\%$, and $2.35\%$ on Digits, Office-10, and PACS datasets, respectively. This finding highlights the ability of our method to generate the generalized prototypes at the inter-domain level, thereby providing the inter-domain knowledge and improving generalization across different domains.

\noindent \textbf{Effect of MixUp on intra-domain prototypes.} In Table~\ref{tab:intra_domain_comparison}, we evaluate the effect of MixUp on our intra-domain prototypes. The results show that by using MixUp at the feature level, our method achieves better generalization than other intra-domain prototype variations. This finding underscores that feature-level MixUp produces richer prototypes, helping to prevent overfitting to specific domains. Additionally, the consistent performance across domains emphasizes the robustness of our feature-level MixUp approach in capturing diverse semantic representations, thereby strengthening the overall model performance under domain shift.
\begin{figure}[]
    \centering
    \begin{subfigure}{0.5\linewidth}
        \centering
        \includegraphics[width=\linewidth]{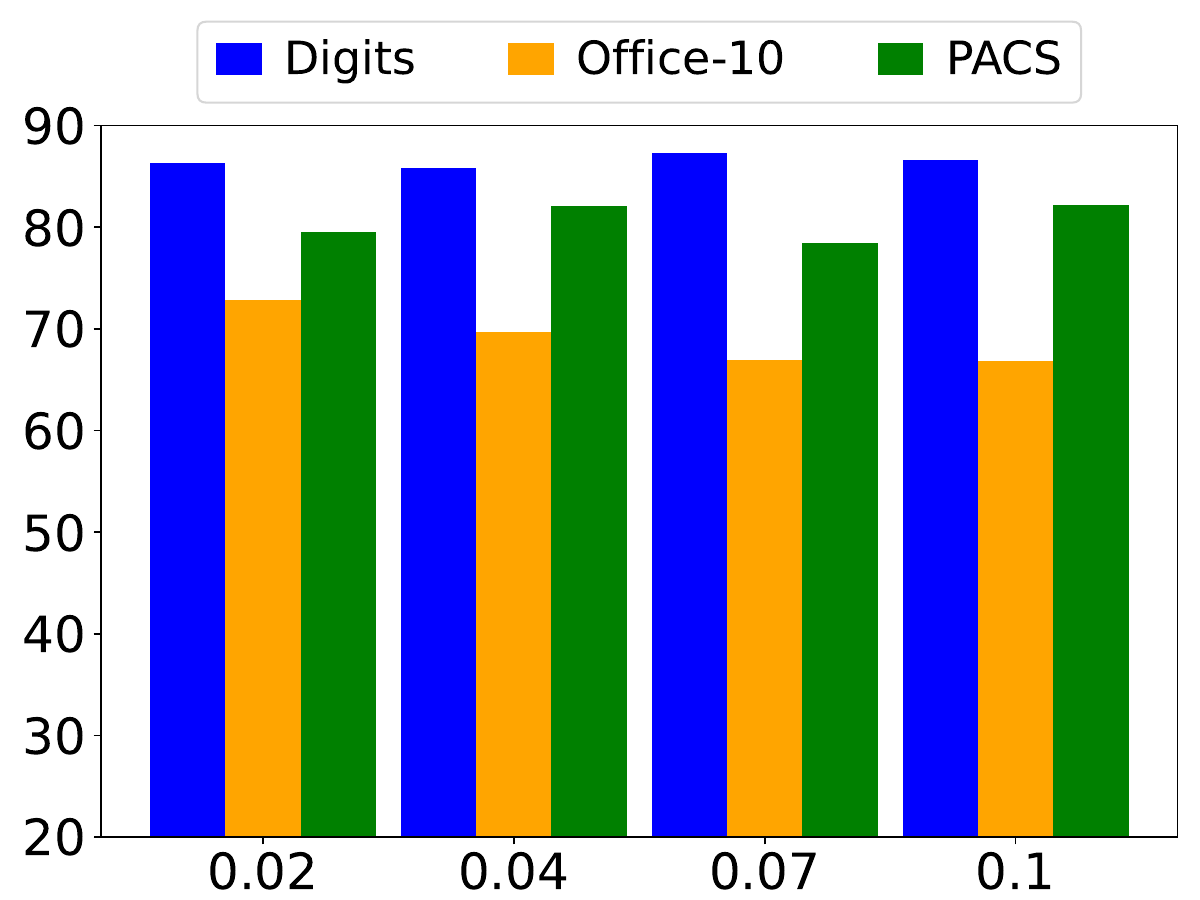}
        \caption{Temperature $\tau$}
        \label{fig:temperature}
    \end{subfigure}%
    \begin{subfigure}{0.5\linewidth}
        \centering
        \includegraphics[width=\linewidth]{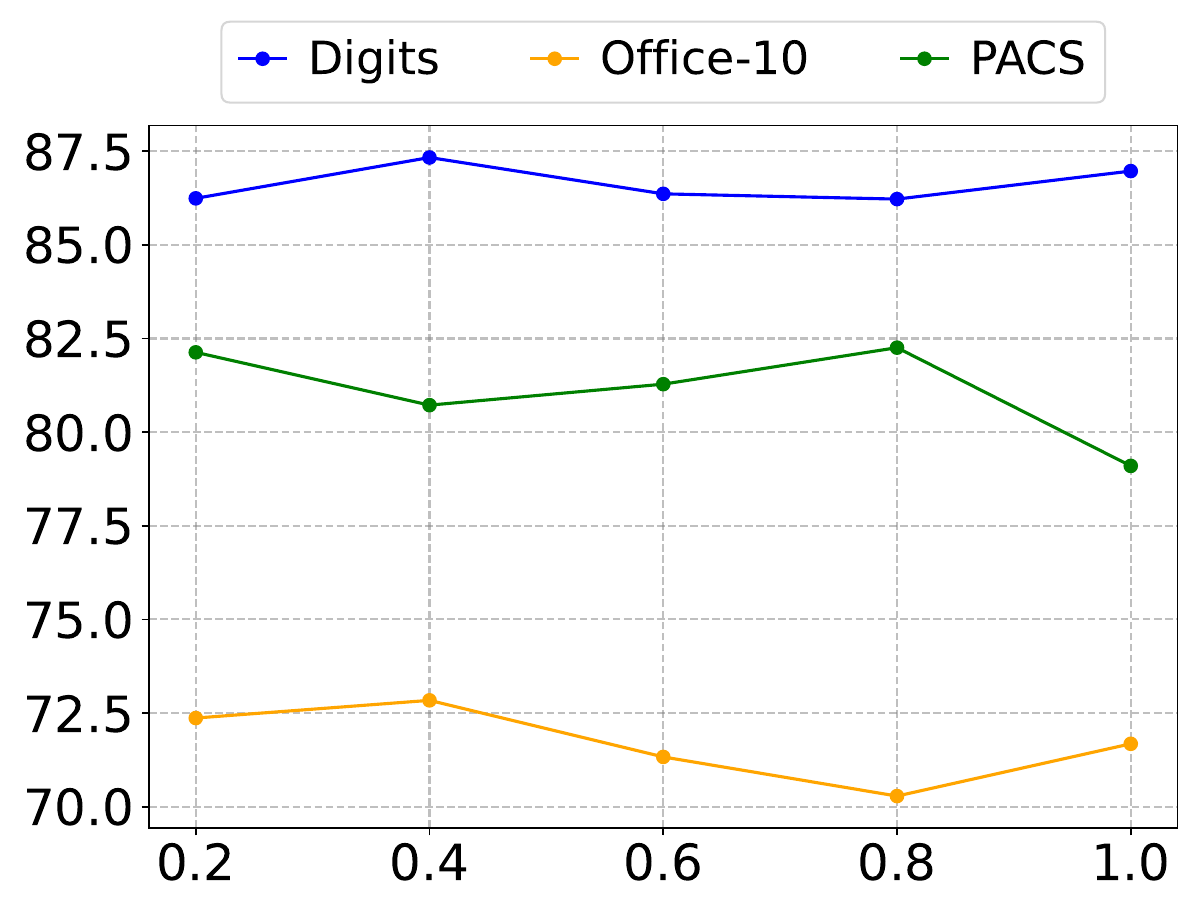}
        \caption{$\alpha$}
        \label{fig:gamma}
    \end{subfigure}
    \caption{Analysis of I$^2$PFL's performance across all datasets with varying values of temperature $\tau$ and $\alpha$ parameters for the Beta distribution. 
    }
    \label{fig:hyper_param}
\end{figure}
\begin{figure}[]
    \centering
    \begin{subfigure}{0.5\linewidth}
        \centering
        \includegraphics[width=\linewidth]{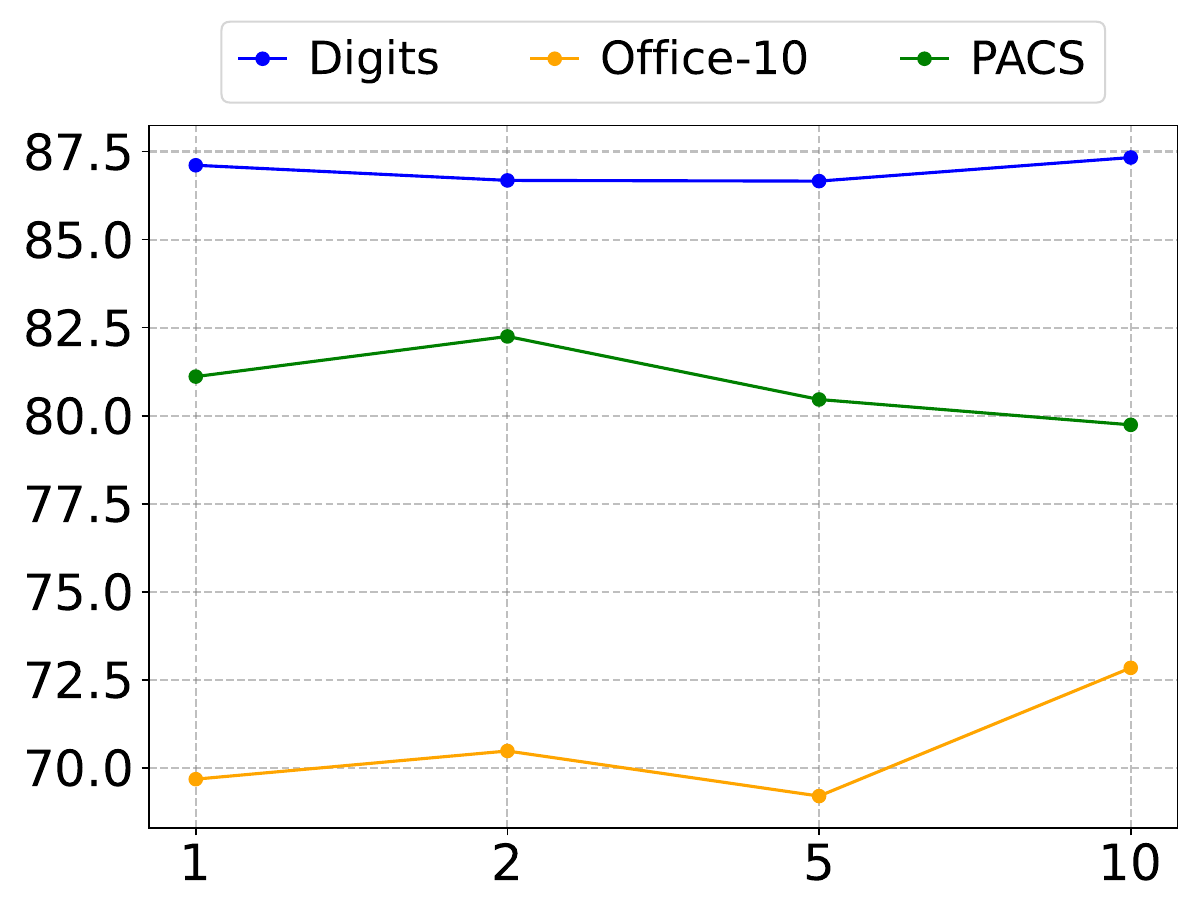}
        \caption{$\lambda_{intra}$}
        \label{fig:intra}
    \end{subfigure}%
    \begin{subfigure}{0.5\linewidth}
        \centering
        \includegraphics[width=\linewidth]{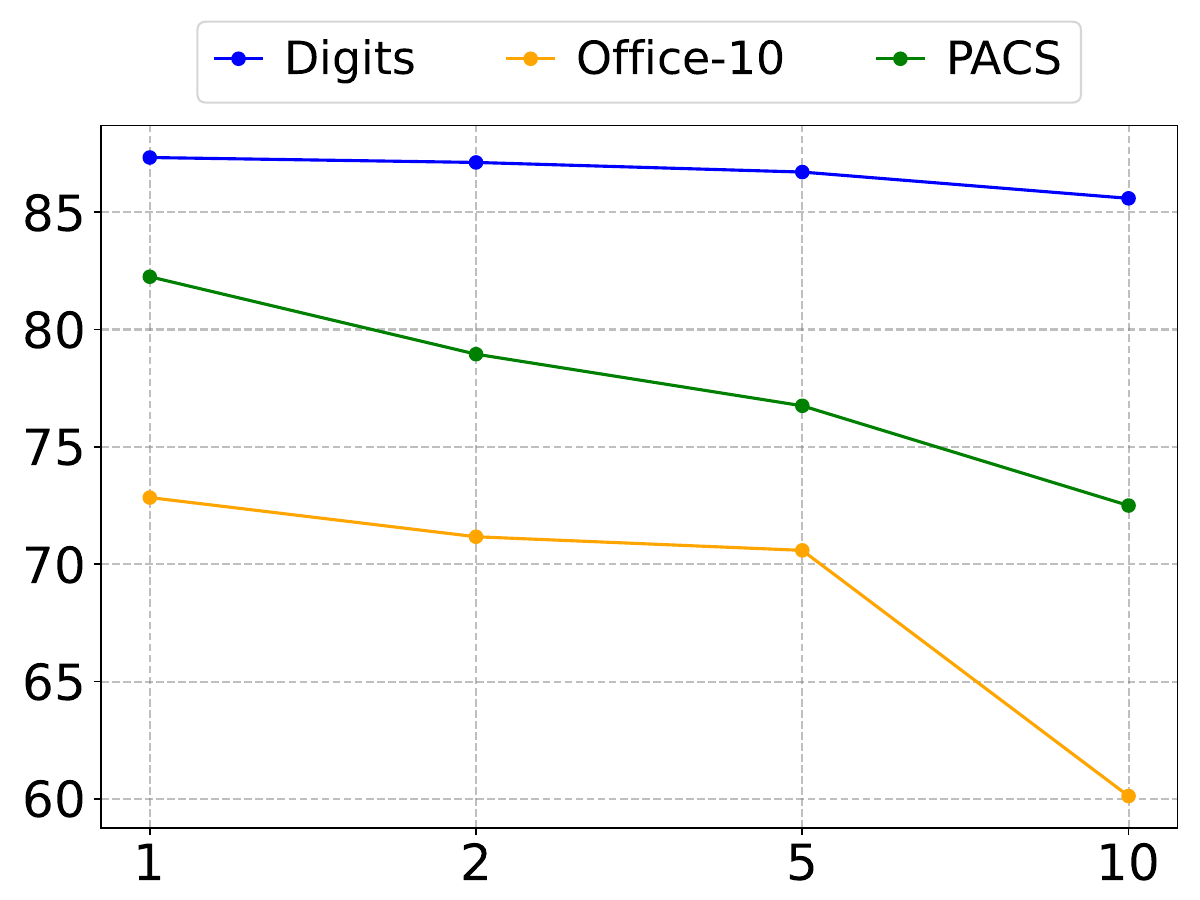}
        \caption{$\lambda_{inter}$}
        \label{fig:inter}
    \end{subfigure}
    \caption{Analysis of I$^2$PFL's performance across all datasets with varying values of $\lambda_{intra}$ and $\lambda_{inter}$. 
    }
    \label{hyper_param_intra_inter}
\end{figure}

\noindent \textbf{Effect of different prototype components.}
We present performance curves in Fig.~\ref{curve_prototypes} to illustrate the impact of different prototype components across all datasets. The results show that intra- and inter-domain prototypes contribute to the convergence of I$^2$PFL, underscoring the effectiveness of combining these prototype types. Notably, intra-domain prototypes substantially impact performance across all datasets, highlighting the benefits of leveraging augmented prototypes locally. These observations confirm the importance of integrating intra- and inter-domain prototypes for optimal performance. 

\noindent \textbf{Privacy preservation analysis.} Following DBE~\cite{zhang2023eliminating}, we added a differential privacy (DP) analysis in different datasets. We added Gaussian noise to client prototypes $\tilde{p}^k_m = p^k_m + q \cdot \mathcal{N}\bigl(0, s^2\bigr)$ with perturbation coefficient $q=0.2$ for the noise and a scale parameter $s=0.05$. As shown in Table~\ref{dp_performance}, the results demonstrate that our method maintains comparable accuracy even under privacy constraints.

\noindent \textbf{Ablation study on various hyper-parameters.} 
We illustrate the performance impact by temperature parameter $\tau$ and $\alpha$ used for Beta distribution in Fig.~\ref{fig:hyper_param}. For the Digits dataset, optimal performance is achieved with $\tau=0.07$ and $\alpha=0.4$. In the Office-10 dataset, the best results are obtained with $\tau=0.02$ and $\alpha=0.4$. In the PACS dataset, optimal performance occurs with $\tau=0.04$ and $\alpha=0.2$. These hyperparameters are used by default in all experiments. We demonstrate the performance impact of the hyperparameters $\lambda_{intra}$ and  $\lambda_{inter}$ in Fig.~\ref{hyper_param_intra_inter}. As shown in the figure, for the Digits and Office-10 datasets, the optimal performance is achieved when $\lambda_{intra}=10$ and $\lambda_{inter}=1$.  In contrast, for the PACS dataset, the best performance occurs with $\lambda_{intra}=2$ and $\lambda_{inter}=1$. These results highlight that increasing the hyperparameter for the intra-domain prototype component ($\lambda_{intra}$) enhances performance, demonstrating the importance of intra-domain prototype alignment in our method.


\section{Conclusion}
This paper introduces I$^2$PFL, a novel prototype-based FL framework designed to mitigate domain shifts in FL. Our approach incorporates two key components: intra-domain prototypes and inter-domain prototypes. Specifically, we introduce the intra-domain prototypes with MixUp-based augmented prototypes. Moreover, we propose a novel prototype reweighting scheme for inter-domain prototypes to generate the generalized prototypes. We use contrastive learning with generalized prototypes to provide inter-domain knowledge and guide local training. Furthermore, we enhance local feature diversity by encouraging alignment between local features and the augmented prototypes. By integrating intra- and inter-domain prototypes, we significantly improve the generalization of the global model and address domain shifts in federated learning. Experiments on three image classification datasets demonstrate the superior performance of IIPFL compared to other state-of-the-art methods.





\bibliographystyle{IEEEtran}
\bibliography{reference}

\begin{thebibliography}{10}
\providecommand{\url}[1]{#1}
\csname url@samestyle\endcsname
\providecommand{\newblock}{\relax}
\providecommand{\bibinfo}[2]{#2}
\providecommand{\BIBentrySTDinterwordspacing}{\spaceskip=0pt\relax}
\providecommand{\BIBentryALTinterwordstretchfactor}{4}
\providecommand{\BIBentryALTinterwordspacing}{\spaceskip=\fontdimen2\font plus
\BIBentryALTinterwordstretchfactor\fontdimen3\font minus \fontdimen4\font\relax}
\providecommand{\BIBforeignlanguage}[2]{{%
\expandafter\ifx\csname l@#1\endcsname\relax
\typeout{** WARNING: IEEEtran.bst: No hyphenation pattern has been}%
\typeout{** loaded for the language `#1'. Using the pattern for}%
\typeout{** the default language instead.}%
\else
\language=\csname l@#1\endcsname
\fi
#2}}
\providecommand{\BIBdecl}{\relax}
\BIBdecl

\bibitem{mcmahan2017communication}
B.~McMahan, E.~Moore, D.~Ramage, S.~Hampson, and B.~A. y~Arcas, ``Communication-efficient learning of deep networks from decentralized data,'' in \emph{Artificial intelligence and statistics}.\hskip 1em plus 0.5em minus 0.4em\relax PMLR, 2017, pp. 1273--1282.

\bibitem{li2020federated}
T.~Li, A.~K. Sahu, M.~Zaheer, M.~Sanjabi, A.~Talwalkar, and V.~Smith, ``Federated optimization in heterogeneous networks,'' \emph{Proceedings of Machine learning and systems}, vol.~2, pp. 429--450, 2020.

\bibitem{li2021survey}
Q.~Li, Z.~Wen, Z.~Wu, S.~Hu, N.~Wang, Y.~Li, X.~Liu, and B.~He, ``A survey on federated learning systems: Vision, hype and reality for data privacy and protection,'' \emph{IEEE Transactions on Knowledge and Data Engineering}, vol.~35, no.~4, pp. 3347--3366, 2021.

\bibitem{rieke2020future}
N.~Rieke, J.~Hancox, W.~Li, F.~Milletari, H.~R. Roth, S.~Albarqouni, S.~Bakas, M.~N. Galtier, B.~A. Landman, K.~Maier-Hein \emph{et~al.}, ``The future of digital health with federated learning,'' \emph{NPJ digital medicine}, vol.~3, no.~1, pp. 1--7, 2020.

\bibitem{kairouz2021advances}
P.~Kairouz, H.~B. McMahan, B.~Avent, A.~Bellet, M.~Bennis, A.~N. Bhagoji, K.~Bonawitz, Z.~Charles, G.~Cormode, R.~Cummings \emph{et~al.}, ``Advances and open problems in federated learning,'' \emph{Foundations and trends{\textregistered} in machine learning}, vol.~14, no. 1--2, pp. 1--210, 2021.

\bibitem{xu2023federated}
Q.~Xu, R.~Zhang, Y.~Zhang, Y.-Y. Wu, and Y.~Wang, ``Federated adversarial domain hallucination for privacy-preserving domain generalization,'' \emph{IEEE Transactions on Multimedia}, vol.~26, pp. 1--14, 2023.

\bibitem{liu2024vertical}
Y.~Liu, Y.~Kang, T.~Zou, Y.~Pu, Y.~He, X.~Ye, Y.~Ouyang, Y.-Q. Zhang, and Q.~Yang, ``Vertical federated learning: Concepts, advances, and challenges,'' \emph{IEEE Transactions on Knowledge and Data Engineering}, 2024.

\bibitem{chen2024think}
J.~Chen, B.~Ma, H.~Cui, and Y.~Xia, ``Think twice before selection: Federated evidential active learning for medical image analysis with domain shifts,'' in \emph{Proceedings of the IEEE/CVF Conference on Computer Vision and Pattern Recognition}, 2024, pp. 11\,439--11\,449.

\bibitem{wang2025federated_tmm}
L.~Wang, S.~Wang, Q.~Zhang, Q.~Wu, and M.~Xu, ``Federated user preference modeling for privacy-preserving cross-domain recommendation,'' \emph{IEEE Transactions on Multimedia}, 2025.

\bibitem{mendieta2022local}
M.~Mendieta, T.~Yang, P.~Wang, M.~Lee, Z.~Ding, and C.~Chen, ``Local learning matters: Rethinking data heterogeneity in federated learning,'' in \emph{Proceedings of the IEEE/CVF Conference on Computer Vision and Pattern Recognition}, 2022, pp. 8397--8406.

\bibitem{li2022federated}
Q.~Li, Y.~Diao, Q.~Chen, and B.~He, ``Federated learning on non-iid data silos: An experimental study,'' in \emph{2022 IEEE 38th international conference on data engineering (ICDE)}.\hskip 1em plus 0.5em minus 0.4em\relax IEEE, 2022, pp. 965--978.

\bibitem{pmlr-v119-karimireddy20a}
\BIBentryALTinterwordspacing
S.~P. Karimireddy, S.~Kale, M.~Mohri, S.~Reddi, S.~Stich, and A.~T. Suresh, ``{SCAFFOLD}: Stochastic controlled averaging for federated learning,'' in \emph{Proceedings of the 37th International Conference on Machine Learning}, ser. Proceedings of Machine Learning Research, H.~D. III and A.~Singh, Eds., vol. 119.\hskip 1em plus 0.5em minus 0.4em\relax PMLR, 13--18 Jul 2020, pp. 5132--5143. [Online]. Available: \url{https://proceedings.mlr.press/v119/karimireddy20a.html}
\BIBentrySTDinterwordspacing

\bibitem{hsu2019measuring}
T.-M.~H. Hsu, H.~Qi, and M.~Brown, ``Measuring the effects of non-identical data distribution for federated visual classification,'' \emph{arXiv preprint arXiv:1909.06335}, 2019.

\bibitem{Wang2020Federated}
\BIBentryALTinterwordspacing
H.~Wang, M.~Yurochkin, Y.~Sun, D.~Papailiopoulos, and Y.~Khazaeni, ``Federated learning with matched averaging,'' in \emph{International Conference on Learning Representations}, 2020. [Online]. Available: \url{https://openreview.net/forum?id=BkluqlSFDS}
\BIBentrySTDinterwordspacing

\bibitem{wang2024aggregation}
Y.~Wang, H.~Fu, R.~Kanagavelu, Q.~Wei, Y.~Liu, and R.~S.~M. Goh, ``An aggregation-free federated learning for tackling data heterogeneity,'' in \emph{Proceedings of the IEEE/CVF Conference on Computer Vision and Pattern Recognition}, 2024, pp. 26\,233--26\,242.

\bibitem{tan2022federated}
Y.~Tan, G.~Long, J.~Ma, L.~Liu, T.~Zhou, and J.~Jiang, ``Federated learning from pre-trained models: A contrastive learning approach,'' \emph{Advances in neural information processing systems}, vol.~35, pp. 19\,332--19\,344, 2022.

\bibitem{huang2023rethinking}
W.~Huang, M.~Ye, Z.~Shi, H.~Li, and B.~Du, ``Rethinking federated learning with domain shift: A prototype view,'' in \emph{2023 IEEE/CVF Conference on Computer Vision and Pattern Recognition (CVPR)}.\hskip 1em plus 0.5em minus 0.4em\relax IEEE, 2023, pp. 16\,312--16\,322.

\bibitem{wu2024prototype}
A.~Wu, J.~Yu, Y.~Wang, and C.~Deng, ``Prototype-decomposed knowledge distillation for learning generalized federated representation,'' \emph{IEEE Transactions on Multimedia}, 2024.

\bibitem{tan2022fedproto}
Y.~Tan, G.~Long, L.~Liu, T.~Zhou, Q.~Lu, J.~Jiang, and C.~Zhang, ``Fedproto: Federated prototype learning across heterogeneous clients,'' in \emph{Proceedings of the AAAI Conference on Artificial Intelligence}, vol.~36, no.~8, 2022, pp. 8432--8440.

\bibitem{wang2024taming}
\BIBentryALTinterwordspacing
L.~Wang, J.~Bian, L.~Zhang, C.~Chen, and J.~Xu, ``Taming cross-domain representation variance in federated prototype learning with heterogeneous data domains,'' in \emph{The Thirty-eighth Annual Conference on Neural Information Processing Systems}, 2024. [Online]. Available: \url{https://openreview.net/forum?id=6SRPizFuaE}
\BIBentrySTDinterwordspacing

\bibitem{zhang2018mixup}
H.~Zhang, M.~Cisse, Y.~N. Dauphin, and D.~Lopez-Paz, ``mixup: Beyond empirical risk minimization,'' in \emph{International Conference on Learning Representations}, 2018.

\bibitem{acar2021federated}
\BIBentryALTinterwordspacing
D.~A.~E. Acar, Y.~Zhao, R.~Matas, M.~Mattina, P.~Whatmough, and V.~Saligrama, ``Federated learning based on dynamic regularization,'' in \emph{International Conference on Learning Representations}, 2021. [Online]. Available: \url{https://openreview.net/forum?id=B7v4QMR6Z9w}
\BIBentrySTDinterwordspacing

\bibitem{t2020personalized}
C.~T~Dinh, N.~Tran, and J.~Nguyen, ``Personalized federated learning with moreau envelopes,'' \emph{Advances in neural information processing systems}, vol.~33, pp. 21\,394--21\,405, 2020.

\bibitem{li2021fedbn}
\BIBentryALTinterwordspacing
X.~Li, M.~JIANG, X.~Zhang, M.~Kamp, and Q.~Dou, ``Fed{BN}: Federated learning on non-{IID} features via local batch normalization,'' in \emph{International Conference on Learning Representations}, 2021. [Online]. Available: \url{https://openreview.net/forum?id=6YEQUn0QICG}
\BIBentrySTDinterwordspacing

\bibitem{zhang2025enhancing}
\BIBentryALTinterwordspacing
J.~Zhang, Y.~Duan, S.~Niu, Y.~CAO, and W.~Y.~B. Lim, ``Enhancing federated domain adaptation with multi-domain prototype-based federated fine-tuning,'' in \emph{The Thirteenth International Conference on Learning Representations}, 2025. [Online]. Available: \url{https://openreview.net/forum?id=3wEGdrV5Cb}
\BIBentrySTDinterwordspacing

\bibitem{wang2025federated}
Z.~Wang, Z.~Wang, X.~Fan, and C.~Wang, ``Federated learning with domain shift eraser,'' in \emph{Proceedings of the Computer Vision and Pattern Recognition Conference}, 2025, pp. 4978--4987.

\bibitem{wu2021collaborative}
G.~Wu and S.~Gong, ``Collaborative optimization and aggregation for decentralized domain generalization and adaptation,'' in \emph{Proceedings of the IEEE/CVF international conference on computer vision}, 2021, pp. 6484--6493.

\bibitem{zhang2023federated}
R.~Zhang, Q.~Xu, J.~Yao, Y.~Zhang, Q.~Tian, and Y.~Wang, ``Federated domain generalization with generalization adjustment,'' in \emph{Proceedings of the IEEE/CVF Conference on Computer Vision and Pattern Recognition}, 2023, pp. 3954--3963.

\bibitem{liu2021feddg}
Q.~Liu, C.~Chen, J.~Qin, Q.~Dou, and P.-A. Heng, ``Feddg: Federated domain generalization on medical image segmentation via episodic learning in continuous frequency space,'' in \emph{Proceedings of the IEEE/CVF conference on computer vision and pattern recognition}, 2021, pp. 1013--1023.

\bibitem{le2024efficiently}
K.~Le, L.~Ho, C.~Do, D.~Le-Phuoc, and K.-S. Wong, ``Efficiently assemble normalization layers and regularization for federated domain generalization,'' in \emph{Proceedings of the IEEE/CVF Conference on Computer Vision and Pattern Recognition}, 2024, pp. 6027--6036.

\bibitem{snell2017prototypical}
J.~Snell, K.~Swersky, and R.~Zemel, ``Prototypical networks for few-shot learning,'' \emph{Advances in neural information processing systems}, vol.~30, 2017.

\bibitem{tian2020rethinking}
Y.~Tian, Y.~Wang, D.~Krishnan, J.~B. Tenenbaum, and P.~Isola, ``Rethinking few-shot image classification: a good embedding is all you need?'' in \emph{Computer Vision--ECCV 2020: 16th European Conference, Glasgow, UK, August 23--28, 2020, Proceedings, Part XIV 16}.\hskip 1em plus 0.5em minus 0.4em\relax Springer, 2020, pp. 266--282.

\bibitem{zhu2023transductive}
H.~Zhu and P.~Koniusz, ``Transductive few-shot learning with prototype-based label propagation by iterative graph refinement,'' in \emph{Proceedings of the IEEE/CVF conference on computer vision and pattern recognition}, 2023, pp. 23\,996--24\,006.

\bibitem{zhang2023prototype}
B.~Zhang, X.~Li, Y.~Ye, and S.~Feng, ``Prototype completion for few-shot learning,'' \emph{IEEE Transactions on Pattern Analysis and Machine Intelligence}, vol.~45, no.~10, pp. 12\,250--12\,268, 2023.

\bibitem{li2021prototypical}
\BIBentryALTinterwordspacing
J.~Li, P.~Zhou, C.~Xiong, and S.~Hoi, ``Prototypical contrastive learning of unsupervised representations,'' in \emph{International Conference on Learning Representations}, 2021. [Online]. Available: \url{https://openreview.net/forum?id=KmykpuSrjcq}
\BIBentrySTDinterwordspacing

\bibitem{gao2023prototype}
K.~Gao, A.~Yu, X.~You, C.~Qiu, and B.~Liu, ``Prototype and context-enhanced learning for unsupervised domain adaptation semantic segmentation of remote sensing images,'' \emph{IEEE Transactions on Geoscience and Remote Sensing}, vol.~61, pp. 1--16, 2023.

\bibitem{cui2024effective}
H.~Cui, L.~Zhao, F.~Li, L.~Zhu, X.~Han, and J.~Li, ``Effective comparative prototype hashing for unsupervised domain adaptation,'' in \emph{Proceedings of the AAAI Conference on Artificial Intelligence}, vol.~38, no.~8, 2024, pp. 8329--8337.

\bibitem{qiao2023mp}
Y.~Qiao, M.~S. Munir, A.~Adhikary, H.~Q. Le, A.~D. Raha, C.~Zhang, and C.~S. Hong, ``Mp-fedcl: Multi-prototype federated contrastive learning for edge intelligence,'' \emph{IEEE Internet of Things journal}, 2023.

\bibitem{fu2025federated}
L.~Fu, S.~Huang, Y.~Lai, C.~Zhang, H.-N. Dai, Z.~Zheng, and C.~Chen, ``Federated domain-independent prototype learning with alignments of representation and parameter spaces for feature shift,'' \emph{IEEE Transactions on Mobile Computing}, 2025.

\bibitem{oord2018representation}
A.~v.~d. Oord, Y.~Li, and O.~Vinyals, ``Representation learning with contrastive predictive coding,'' \emph{arXiv preprint arXiv:1807.03748}, 2018.

\bibitem{mu2023fedproc}
X.~Mu, Y.~Shen, K.~Cheng, X.~Geng, J.~Fu, T.~Zhang, and Z.~Zhang, ``Fedproc: Prototypical contrastive federated learning on non-iid data,'' \emph{Future Generation Computer Systems}, vol. 143, pp. 93--104, 2023.

\bibitem{hull1994database}
J.~J. Hull, ``A database for handwritten text recognition research,'' \emph{IEEE Transactions on pattern analysis and machine intelligence}, vol.~16, no.~5, pp. 550--554, 1994.

\bibitem{lecun1998gradient}
Y.~LeCun, L.~Bottou, Y.~Bengio, and P.~Haffner, ``Gradient-based learning applied to document recognition,'' \emph{Proceedings of the IEEE}, vol.~86, no.~11, pp. 2278--2324, 1998.

\bibitem{netzer2011reading}
Y.~Netzer, T.~Wang, A.~Coates, A.~Bissacco, B.~Wu, A.~Y. Ng \emph{et~al.}, ``Reading digits in natural images with unsupervised feature learning,'' in \emph{NIPS workshop on deep learning and unsupervised feature learning}, vol. 2011, no.~2.\hskip 1em plus 0.5em minus 0.4em\relax Granada, 2011, p.~4.

\bibitem{roy2018effects}
P.~Roy, S.~Ghosh, S.~Bhattacharya, and U.~Pal, ``Effects of degradations on deep neural network architectures,'' \emph{arXiv preprint arXiv:1807.10108}, 2018.

\bibitem{gong2012geodesic}
B.~Gong, Y.~Shi, F.~Sha, and K.~Grauman, ``Geodesic flow kernel for unsupervised domain adaptation,'' in \emph{2012 IEEE conference on computer vision and pattern recognition}.\hskip 1em plus 0.5em minus 0.4em\relax IEEE, 2012, pp. 2066--2073.

\bibitem{li2017deeper}
D.~Li, Y.~Yang, Y.-Z. Song, and T.~M. Hospedales, ``Deeper, broader and artier domain generalization,'' in \emph{Proceedings of the IEEE international conference on computer vision}, 2017, pp. 5542--5550.

\bibitem{li2021model}
Q.~Li, B.~He, and D.~Song, ``Model-contrastive federated learning,'' in \emph{Proceedings of the IEEE/CVF conference on computer vision and pattern recognition}, 2021, pp. 10\,713--10\,722.

\bibitem{he2016deep}
K.~He, X.~Zhang, S.~Ren, and J.~Sun, ``Deep residual learning for image recognition,'' in \emph{Proceedings of the IEEE conference on computer vision and pattern recognition}, 2016, pp. 770--778.

\bibitem{robbins1951stochastic}
H.~Robbins and S.~Monro, ``A stochastic approximation method,'' \emph{The annals of mathematical statistics}, pp. 400--407, 1951.

\bibitem{van2008visualizing}
L.~Van~der Maaten and G.~Hinton, ``Visualizing data using t-sne.'' \emph{Journal of machine learning research}, vol.~9, no.~11, 2008.

\bibitem{sarfraz2019efficient}
S.~Sarfraz, V.~Sharma, and R.~Stiefelhagen, ``Efficient parameter-free clustering using first neighbor relations,'' in \emph{Proceedings of the IEEE/CVF conference on computer vision and pattern recognition}, 2019, pp. 8934--8943.

\bibitem{zhang2023eliminating}
J.~Zhang, Y.~Hua, J.~Cao, H.~Wang, T.~Song, Z.~Xue, R.~Ma, and H.~Guan, ``Eliminating domain bias for federated learning in representation space.'' in \emph{NeurIPS}, 2023.

\end{thebibliography}

\end{document}